\documentclass{article}
\usepackage[super,sort&compress,comma]{natbib}

\usepackage[english]{babel}

\usepackage[letterpaper,top=2cm,bottom=2cm,left=3cm,right=3cm,marginparwidth=1.75cm]{geometry}

\usepackage{array, makecell, booktabs, tabularx, multirow}

\usepackage{siunitx}
\usepackage{ragged2e}

\usepackage{amsmath, amssymb}

\usepackage{graphicx}
\usepackage{caption}
\usepackage{float}

\makeatletter
\renewcommand\@makefntext[1]{%
  \setlength{\parindent}{0pt}
  \setlength{\leftskip}{0pt}
  \noindent
  \makebox[1em][r]{\@makefnmark\ }%
  #1%
}
\renewcommand\thanks[1]{%
  \footnotemark
  \protected@xdef\@thanks{\@thanks
    \protect\footnotetext[\the\c@footnote]{\noindent #1}%
  }%
}
\makeatother

\usepackage[colorlinks=true, allcolors=blue]{hyperref}

\title{UrbanPulse: A Cross-City Deep Learning Framework for Ultra-Fine-Grained Population Transfer Prediction}
\author{
  Hongrong Yang\thanks{\footnotesize\textit{Department of Civil Engineering and Engineering Mechanics, Fu Foundation School of Engineering and Applied Science, Columbia University, New York, NY, USA.}}
  \ \  Markus Schläpfer\footnotemark[1]
}
\date{}

\begin{document}
\maketitle

\begin{abstract}

Accurate population flow prediction is essential for urban planning, transportation management, and public health. Yet existing methods face key limitations: traditional models rely on static spatial assumptions, deep learning models struggle with cross-city generalization, and Large Language Models (LLMs) incur high computational costs while failing to capture spatial structure. Moreover, many approaches sacrifice resolution by clustering Points of Interest (POIs) or restricting coverage to subregions, limiting their utility for city-wide analytics. We introduce UrbanPulse, a scalable deep learning framework that delivers ultra-fine-grained, city-wide OD flow predictions by treating each POI as an individual node. It combines a temporal graph convolutional encoder with a transformer-based decoder to model multi-scale spatiotemporal dependencies. To ensure robust generalization across urban contexts, UrbanPulse employs a three-stage transfer learning strategy: pretraining on large-scale urban graphs, cold-start adaptation, and reinforcement learning fine-tuning.Evaluated on over 103 million cleaned GPS records from three metropolitan areas in California, UrbanPulse achieves state-of-the-art accuracy and scalability. Through efficient transfer learning, UrbanPulse takes a key step toward making high-resolution, AI-powered urban forecasting deployable in practice across diverse cities.

\end{abstract}

\section{Main}
Accurate origin-destination (OD) flow prediction is critical for modern urban systems, supporting data-driven decisions in transportation planning \cite{rong2024interdisciplinary}, infrastructure deployment \cite{sayarshad2024optimization}, and mobility service optimization \cite{tang2024origin}. As cities grow increasingly complex and dynamic, there is rising demand for models capable of forecasting mobility at high spatiotemporal resolution \cite{cabanas2025human,guo2025unifying}. Since mobility unfolds at the level of individual Points of Interest (POIs), each with distinct temporal flow patterns \cite{gallotti2024distorted}, capturing such fine-grained dynamics is essential for tasks like demand-responsive transit and energy-aware infrastructure control \cite{tan2025spatiotemporal}. However, collecting reliable OD matrices remains costly, labor-intensive, and constrained by privacy and coverage limitations—particularly in cities lacking sensing infrastructure or featuring diverse transportation modes \cite{rong2023goddag}. These challenges are further compounded by wide variation in urban form and behavior, which often prevents models trained in one city from generalizing to another \cite{shao2025cross}. Addressing these limitations requires predictive frameworks that combine POI-level granularity with city-scale adaptability, without relying on extensive local data.

Despite the growing interest in fine-grained urban mobility forecasting, existing approaches face fundamental limitations in their ability to deliver fine-grained, scalable, and transferable OD flow predictions. Traditional models, such as the gravity and radiation formulations, have long been used to estimate mobility flows. The gravity model, rooted in early spatial interaction studies, assumes that flow between two locations increases with their population and decreases with distance \cite{zipf1946p}. In contrast, the radiation model assumes that flow depends on the population of the origin and destination, moderated by the population within a radius between them \cite{simini2012universal}. While both models are interpretable due to their simple formulations, they rely on fixed spatial partitions and strong equilibrium assumptions \cite{Alessandretti2020,Barbosa2018}, which struggle to capture the dynamic, asymmetric, and temporally evolving nature of population flows in modern cities. Moreover, they cannot scale to POI-level predictions without substantial loss of accuracy \cite{Schlapfer2021,Song2010}.

Deep learning has markedly advanced urban mobility forecasting by learning nonlinear spatiotemporal dependencies from large-scale movement data \cite{luca2021survey}. Early grid-based approaches \cite{ke2018hexagon,zhang2017deep,zhang2019flow,lin2019deepstn}, which combined convolutional neural networks with time-series forecasting models like Long Short-Term Memory \cite{hochreiter1997long} or Recurrent Neural Network \cite{elman1990finding}, struggled with edge effects and coarse spatial resolution, limiting their ability to capture fine-grained POI-level dynamics. To overcome these limitations, topological methods, particularly graph convolutional networks (GCNs)-based methods \cite{ye2021coupled,liu2023gnn,xue2022quantifying,lee2022ddp}, were introduced to represent urban mobility as graphs over POIs or road networks, incorporating temporal components. While effective in modeling non-Euclidean spatial structure, such methods rely on static adjacency matrices that fail to adapt to dynamic urban changes and generalize poorly across cities \cite{pareja2020evolvegcn}. Spatiotemporal graph neural networks methods \cite{zhao2019t,wang2021gsnet,jin2023spatio,wang2024spatiotemporal} build on this framework by jointly modeling spatial and temporal dependencies, yet face scalability constraints \cite{cini2023scalable} in large urban systems and require retraining for each new city due to their dependence on local structural features \cite{pareja2020evolvegcn}. To address cross-city deployment challenges, transfer learning approaches \cite{wang2019cross,jin2022selective} have emerged as a promising strategy forreducing reliance on labeled data in target cities. Transformer-based large language models (LLMs) \cite{li2024urbangpt,liu2024spatial,liang2024exploring,li2024chatsumo} have also been explored for mobility forecasting tasks in urban science due to their capacity to capture long-range temporal dependencies. However, LLMs treat mobility data as flat sequences \cite{tan2024language}, ignoring spatial structure and failing to model localized POI-level flows \cite{manvi2024geollm} . Their substantial computational demands \cite{zeng2023transformers} and dependence on extensive pre-training \cite{chang2024large} further limit their practicality for real-time, fine-grained forecasting.

Our work bridges a fundamental divide in urban mobility forecasting through a hybrid architecture that combines temporal graph convolutions with transformer-based decoding, enabling both fine-grained prediction accuracy and cross-city generalizability. The temporal graph component captures localized mobility patterns, while the transformer decoder models long-range urban dependencies - together forming a powerful framework for spatial-temporal forecasting. Building upon this architectural foundation, we introduce a three-stage transfer learning approach that systematically converts universal mobility knowledge into city-specific forecasting capabilities: pretraining establishes core patterns, cold-start adaptation preserves these fundamentals while adjusting to new urban layouts, and reinforcement learning fine-tunes local predictive accuracy.
 
The results not only establish new state-of-the-art performance metrics but, more importantly, point toward a future where accurate urban mobility forecasting can be rapidly deployed in new cities- significantly reducing the need for costly data collection and high computational of retraining procedures. This represents a significant step forward in making AI-powered urban analytics truly scalable and accessible across diverse geographical contexts.

\section{Results}
We present three key results: the generalization performance of the pre-trained model on the Los Angeles dataset, cross-city transfer evaluation with fine-tuning on the San Francisco dataset, and ablation results examining the impact of architectural and training components.

\subsection{General Performance on Los Angeles}

\begin{figure}[htbp]
\centering
\includegraphics[width=0.95\linewidth]{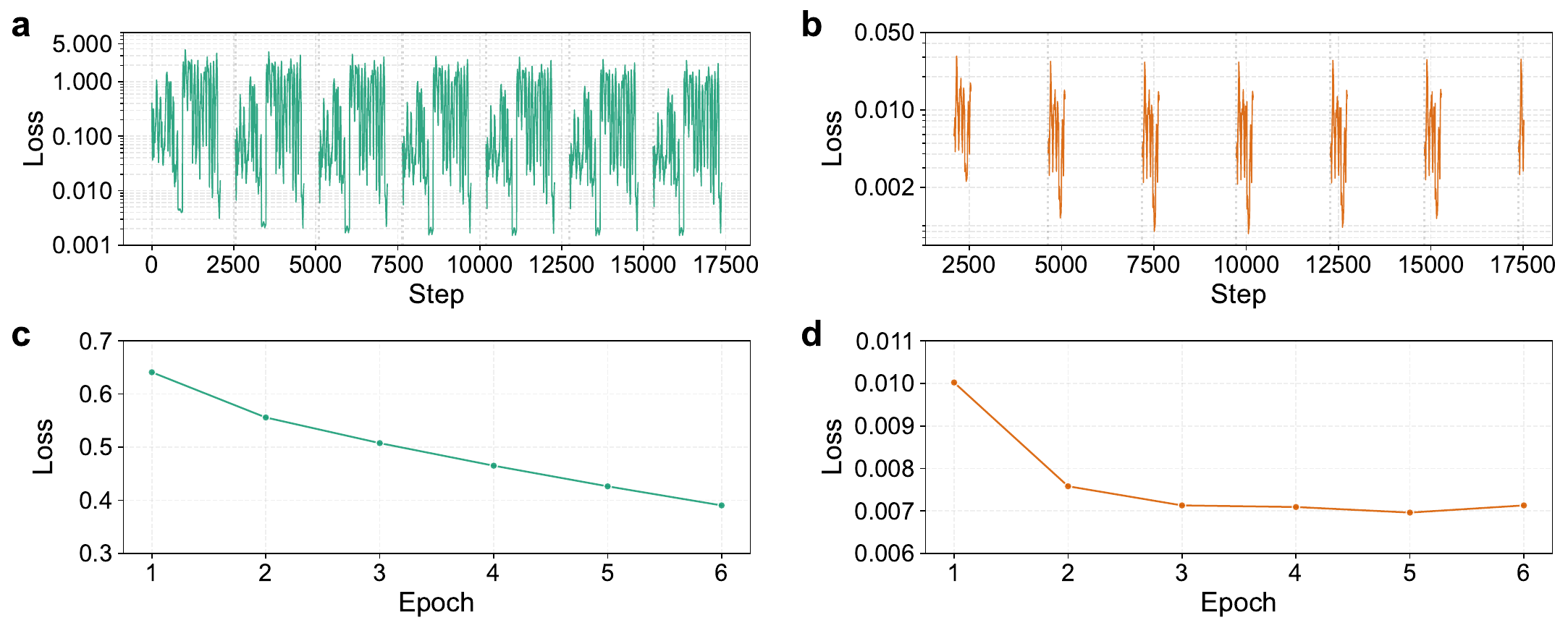}
\caption*{\textbf{Fig. 1 \textbar{} Training and validation loss dynamics of the pre-trained model on the Los Angeles dataset.} \textbf{a},\textbf{b}, Step-wise loss during model pre-training. \textbf{a}, Log-scaled training loss decreases consistently. \textbf{b}, Validation loss shows greater fluctuation but follows a converging trend.
\textbf{c}, \textbf{d}, Epoch-level loss over the full training period. \textbf{c}, Training loss gradually decreases over epochs. \textbf{d}, Validation loss reaches a minimum of 0.0104, indicating good generalization to unseen data.}
\label{fig:train_val_loss}
\end{figure}

We train the basic model on large-scale urban mobility datasets collected from Los Angeles, using fine-grained GPS mobility traces aggregated into 15-minute intervals. Each city is represented as a dynamic spatiotemporal graph, where nodes correspond to real-world points of interest (POIs) extracted from OpenStreetMap, and directed edges capture observed transitions between POIs. Node features include both static attributes—such as POI type and dynamic signals, including population counts, geographic coordinates (latitude and longitude) and weather conditions (temperature, precipitation, and wind speed), the latter derived from the ERA5 reanalysis dataset provided by the Copernicus Climate Data Store.

All real-valued features—including population counts, geographic coordinates, and weather variables—are normalized to the range [0,1] using min–max scaling based on training data statistics. Categorical POI types are embedded into a low-dimensional space to capture functional similarities across locations. To mitigate the impact of outliers, precipitation is first capped at 50 mm/h and log-transformed prior to normalization. The model is trained to predict edge-level flows across the Los Angeles POI network of 38,628 nodes over 12 consecutive 15-minute intervals using a sliding window, producing a continuous 3-hour forecast from a fixed starting point without iterative reforecasting. This setup allows the model to learn high-resolution population flow dynamics under varying spatial and environmental conditions. 

We monitor the training dynamics of the pre-trained model on the Los Angeles dataset (Fig. 1a, c) and select the checkpoint with the lowest validation loss (Fig. 1b, d) to ensure optimal generalization. To illustrate the model’s fine-grained spatiotemporal capabilities, Fig. 2a visualizes population flows at a single 15-minute interval, revealing its ability to capture structured movement patterns across the urban network. Edges are color-coded by prediction outcome—green for correct, red for overpredicted, and blue for underpredicted—highlighting spatial accuracy. At this resolution, the model achieves near-perfect spatial fidelity, with 99.79\% of edges predicted correctly and errors confined to isolated network segments. This precision is further quantified in Fig. 2b, where temporal tracking of prediction outcomes confirms consistent reliability across the entire forecasting window.

The Extended Data figures provide comprehensive regime-specific analyses, with each period's complete 3-hour dynamics visualized through 12 consecutive spatial subgraphs (Extended Data Figs. 1a-3a). These reveal how prediction accuracy evolves across the urban fabric, maintaining the same intuitive color coding while highlighting regime-dependent spatial patterns. Corresponding temporal analyses (Extended Data Figs. 1b-3b) track performance fluctuations across intervals, while cumulative distributions (Extended Data Figs. 1c-3c) provide compact statistical summaries of each period's overall performance on logarithmic scales.

Rigorous evaluation across distinct mobility regimes demonstrates the model's remarkable adaptability. Morning commuter flows (07:00-10:00), characterized by strong unidirectional biases, are predicted with 99.79\% accuracy and remarkably balanced error distribution (41 overpredictions vs. 43 underpredictions). The midday phase (12:00-15:00) reveals marginally lower but still exceptional 99.44\% accuracy, with a slight tendency toward conservative underpredictions (106 vs. 55 overpredictions) during low-volatility conditions. The evening period (17:00-20:00) presents the most significant challenges, with accuracy dipping to 99.39\%. Notably, these evening errors exhibit spatial persistence at specific nodes, particularly in the densest POI areas, recurring across multiple 15-minute intervals (Extended Data Figs. 3a).

The performance comparison evaluates Urban Pulse against five state-of-the-art baselines: Graph WaveNet \cite{wu2019graph}, EGAT \cite{wang2021egat}, DCRNN \cite{li2017diffusion}. UrbanPulse demonstrates superior performance compared to existing methods, achieving the lowest MSE (1.79) and MAE (2.08) on the Los Angeles test set. It significantly outperforms Graph WaveNet (MSE: 2.37, MAE: 2.44) by approximately 24.47\% and 14.75\% respectively, while showing even greater improvements over EGAT (27.23\% lower MSE) and DCRNN (48.86\% lower MSE). The performance advantages of UrbanPulse stem from several key architectural innovations: (1) The dynamic graph convolution module, which learns time-varying adjacency matrices, enables adaptation to real-time changes in urban connectivity, yielding significant improvements over static-graph approaches like DCRNN. (2) The combined use of sinusoidal encodings and temporal convolutions enhances temporal pattern recognition compared to EGAT’s attention mechanism, likely due to the complementary strengths of modeling periodic and local dependencies, contributing to a 20.91\% MAE reduction. (3) The edge-based Transformer decoder precisely models directional flow relationships, outperforming Graph WaveNet’s node-centric representations, which aggregate edge information indirectly. Together, these components address key limitations in existing methods’ ability to capture multi-scale spatial and temporal dependencies, as demonstrated on the Los Angeles test set, with further evaluations underway to confirm generalizability across diverse urban datasets.

\begin{figure}[t]
\centering
\includegraphics[width=0.95\linewidth]{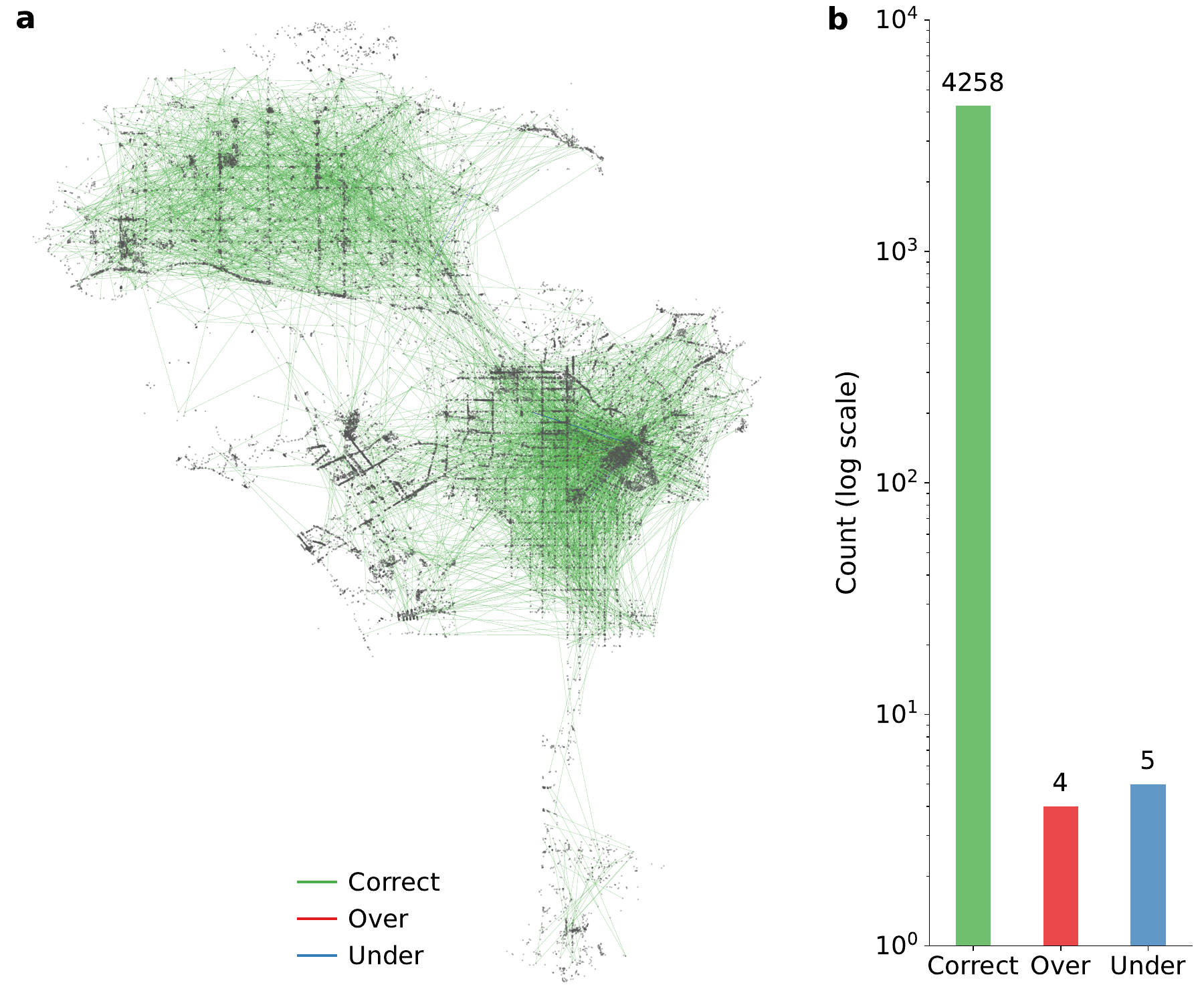}
\caption*{\textbf{Fig. 2 \textbar{} Prediction results of urban population flow at a single 15-minute interval in the Los Angeles test dataset. (07:00, March 29, 2023).} 
\textbf{a}, Visualization of the predicted population transfer network at 07:00 in Los Angeles. 
Nodes represent points of interest with size of 38628. 
Edges indicate predicted population flows, color-coded by prediction error relative to ground truth: correct (green), overpredicted (red), and underpredicted (blue). 
\textbf{b}, Summary of edge-level prediction counts at this time step, plotted on a logarithmic scale.}
\label{fig:prediction_LA}
\end{figure}

\begin{table}[t]
\centering
\caption*{\textbf{Table 1 \textbar{} Performance comparison of different prediction methods on the Los Angeles test set}}
\label{tab:performance_comparison}
\resizebox{\textwidth}{!}{%
\begin{tabular}{
>{\centering\arraybackslash}m{2.0cm}  
*{2}{>{\centering\arraybackslash}m{3.5cm}}  
*{2}{>{\centering\arraybackslash}m{3.0cm}} 
}
\toprule
\textbf{Metric} & \textbf{Urban Pulse} & \textbf{Graph WaveNet} & \textbf{EGAT} & \textbf{DCRNN} \\
\midrule
MSE & \textbf{1.79} & \underline{2.37} & 2.46 & 3.50 \\
MAE & \textbf{2.08} & \underline{2.44} & 2.63 & 3.89 \\
\bottomrule
\addlinespace[4pt]
\end{tabular}%
}
\vspace{4pt}
\noindent\footnotesize
\raggedright
\parbox{\textwidth}{
Note: MSE and MAE values are multiplied by $10^{-2}$ for readability. Bold indicates the best result; underline indicates the second best.
}
\end{table}

\subsection{Cross-City Adaptation on San Francisco}

We employ Los Angeles as the representative mega-city for pretraining, testing transferability to San Francisco to evaluate how well metro-scale patterns generalize to smaller but structurally similar coastal cities, despite their population density differences. The model demonstrates consistent transfer learning capabilities when adapted to San Francisco, as evidenced by the reinforcement learning (RL) training dynamics (Fig. 1). While the raw rewards show expected volatility during exploration (Fig. 3a), the smoothed trajectory reveals consistent performance improvement, with reward values stabilizing after approximately 21000 training episodes (Fig. 3b). This suggests effective policy convergence despite the distinct urban topology of San Francisco.

\begin{figure}[t]
\centering
\includegraphics[width=0.95\linewidth]{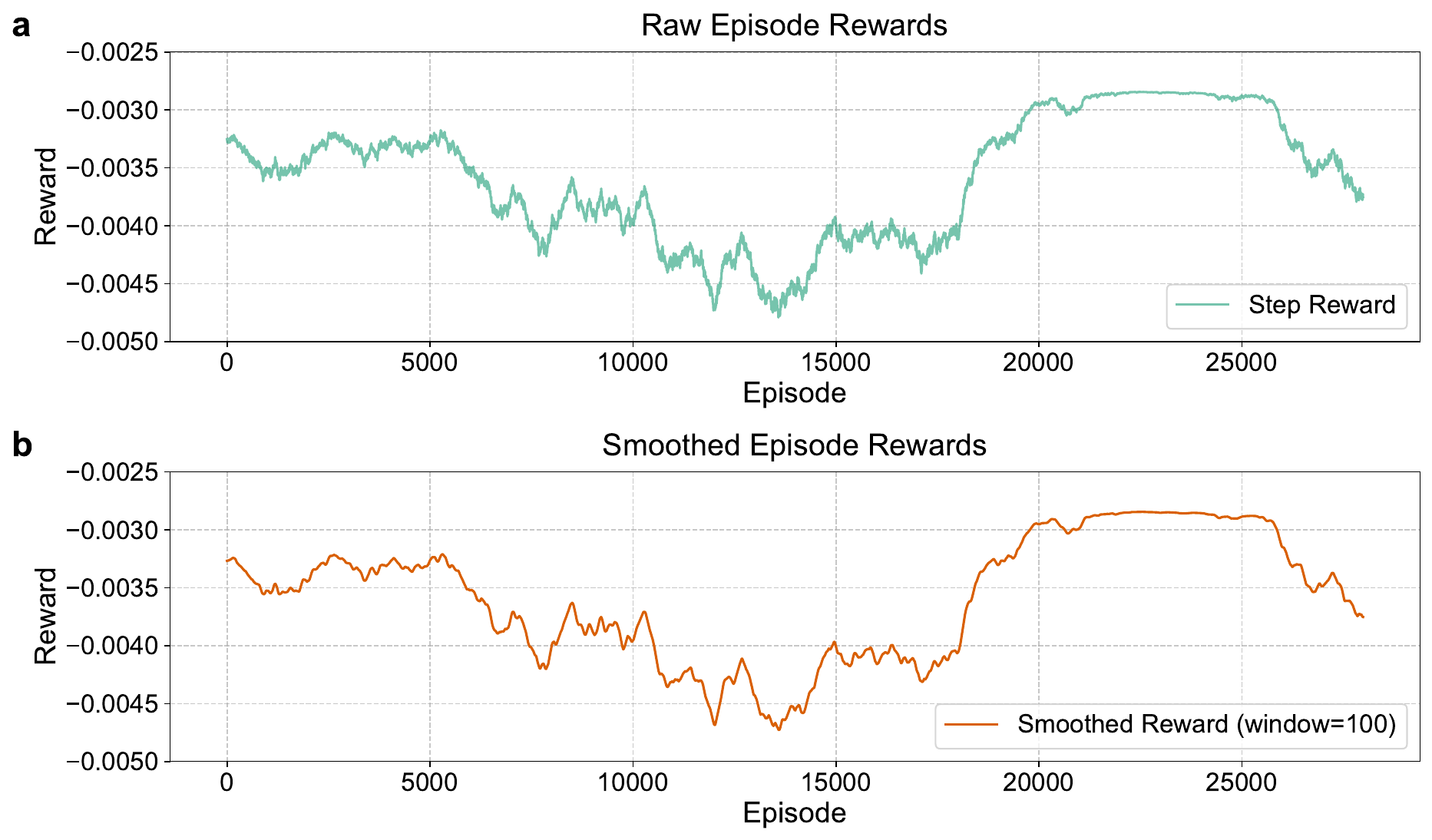}
\caption*{\textbf{Fig. 3 \textbar{} Reinforcement learning (RL) training performance during fine-tuning on San Francisco dataset.} 
\textbf{a}, Raw step-wise rewards for the RL agent during the training period, showing fluctuations in reward values. \textbf{b}, Smoothed rewards computed using a moving average (window size = 100), highlighting the underlying performance trend.}
\label{fig:rl_training}
\end{figure}

The model's spatial prediction accuracy is clearly illustrated in Fig. 4a, showing 99.87\% edge-level accuracy at the 07:00 timepoint. This high precision is sustained throughout the morning commute, as shown in Extended Data Figs. 1b-c, with 99.88\% accuracy and a slight increase in error frequency during the 07:00-08:00 surge (Fig. 4b). The spatial consistency of these patterns is further validated by the full morning sequence in Extended Data Fig. 4a, where error locations do not exhibit persistent clustering across consecutive 15-minute intervals.

Performance remains strong during midday with 99.78\% accuracy, though there is a slight increase in underpredictions compared to the morning. The evening period shows the most pronounced error pattern: while achieving 99.67\% accuracy, it accounts for all 10 overpredictions and 77.4\% of underpredictions. Additionally, when examining error scaling with flow volume, we find that evening periods generate approximately 2.1 times more predictions than midday, but produce 3.1 times more errors. This disproportionate increase and concentration of errors during peak evening hours suggest the model faces greater challenges in predicting high-volume flow periods in San Francisco's mobility patterns.

\begin{figure}[htbp]
\centering
\includegraphics[width=0.95\linewidth]{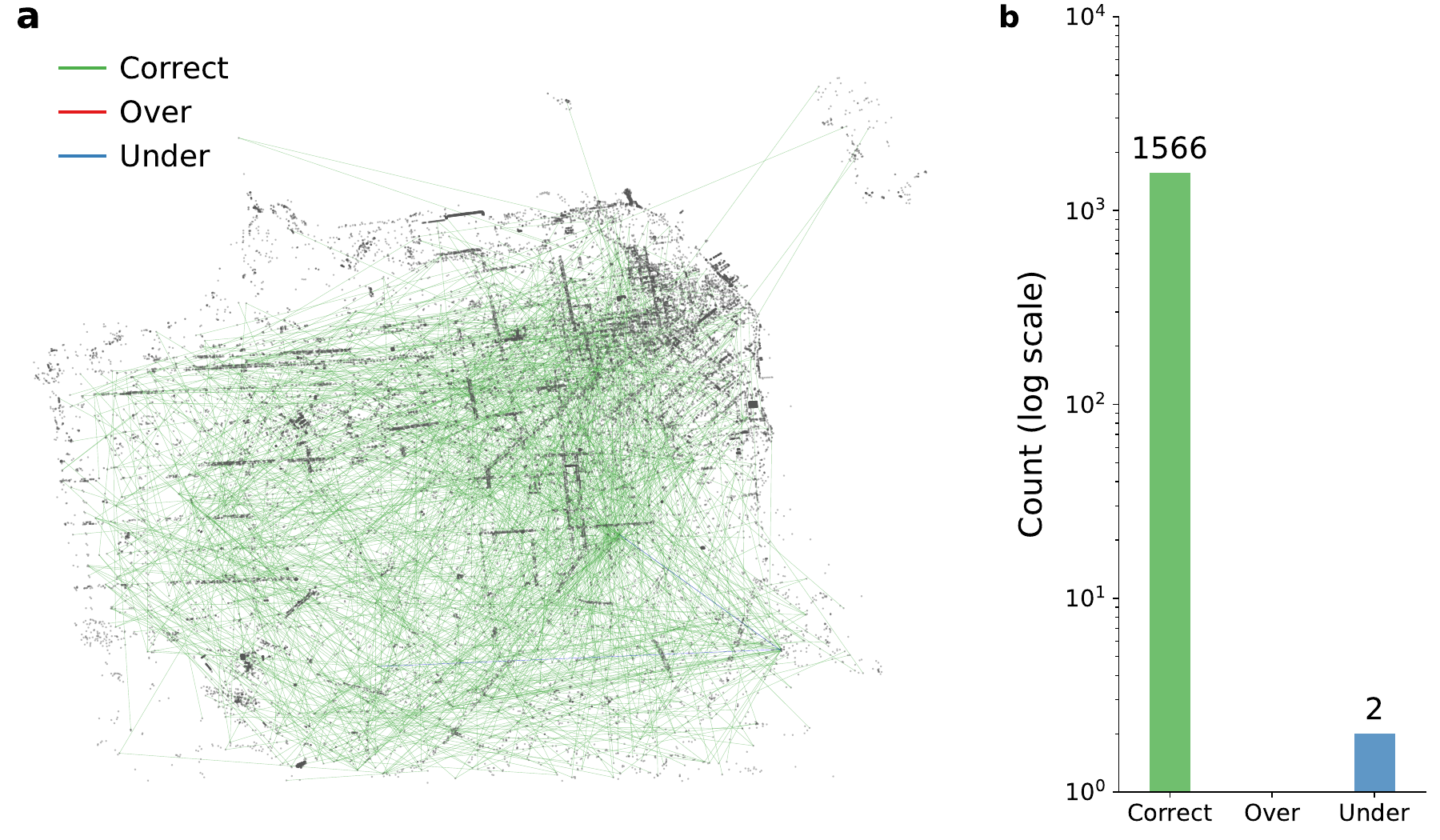}
\caption*{\textbf{Fig. 4 \textbar{} Prediction results of urban population flow at a single 15-minute interval in the San Francisco test dataset. (07:00, March 29, 2023).} 
\textbf{a}, Visualization of the predicted population transfer network at 07:00 in San Francisco. 
Nodes represent points of interest with size of 28580. 
Edges indicate predicted population flows, color-coded by prediction error relative to ground truth: correct (green), overpredicted (red), and underpredicted (blue). 
\textbf{b}, Summary of edge-level prediction counts at this time step, plotted on a logarithmic scale.}
\label{fig:prediction_SF}
\end{figure}

\begin{table}[t]
\centering
\caption*{\textbf{Table 2 \textbar{} Cross-city performance comparison on the San Francisco test set}}
\label{tab:cross_city_comparison}
\resizebox{\textwidth}{!}{%
\begin{tabular}{
>{\centering\arraybackslash}m{2.4cm}  
*{1}{>{\centering\arraybackslash}m{2.6cm}} 
*{2}{>{\centering\arraybackslash}m{2.3cm}} 
*{1}{>{\centering\arraybackslash}m{2.6cm}} 
*{1}{>{\centering\arraybackslash}m{3.2cm}} 
}
\toprule
\textbf{Metric} & \textbf{RL\&CS} & \textbf{SL} & \textbf{CS} & \textbf{SL\&CS} & \textbf{LA Model on SF} \\
\midrule
MSE & \textbf{0.64} & 1.05 & 0.68 & 1.32 & 0.72 \\
MAE & \textbf{1.08} & 2.74 & 1.13 & 1.39 & 2.03 \\
\bottomrule
\addlinespace[4pt]
\end{tabular}%
}
\vspace{4pt}
\noindent\footnotesize
\raggedright
\parbox{\textwidth}{
Note: LA = Los Angeles, SF = San Francisco. CS = Cold Start. SL = Supervised Learning. ``SL \& CS" denotes supervised learning fine-tuning with cold start initialization. ``RL \& CS" denotes reinforcement learning fine-tuning with cold start initialization. MSE and MAE values are multiplied by $10^{-2}$ for readability. Bold values indicate best performance.
}
\end{table}

The cross-city evaluation reveals nuanced dynamics in UrbanPulse's transfer learning framework. The pre-trained Los Angeles model consistently outperforms both the supervised learning (SL) and SL cold-start (CS) models, indicating robust generalization. However, fine-tuning on limited San Francisco (SF) data may lead to slight overfitting or loss of prior knowledge. While the model demonstrates strong zero-shot prediction capability, it still underperforms compared to the CS and CS Reinforcement Learning (RL) approaches—underscoring the need for adaptation to city-specific traffic patterns and localized POI interactions.

The CS model outperforms both SL and SL\&CS approaches, likely due to its more restrained fine-tuning—updating only the deep layers. This strategy preserves the pretrained model’s generalizable spatiotemporal features while avoiding overfitting. The staircase-style GCN design, where later layers have more neurons, ensures that earlier layers capture lower-level features. If the parameters of these layers are well-optimized, freezing them helps maintain their feature-learning capabilities. In contrast, the SL\&CS model exhibits poorer performance, indicating catastrophic forgetting: full fine-tuning corrupts pretrained representations when initialized from cold-start weights. The SL model shows a lower MSE but higher MAE than SL\&CS, suggesting it tolerates frequent small errors but struggles with large deviations.

The reinforcement learning with cold-start initialization (RL\&CS) model’s enhanced performance builds upon the cold-start foundation by integrating RL’s unique optimization dynamics. Unlike supervised learning, which minimizes errors directly, RL utilizes a temporal credit assignment mechanism that decomposes prediction accuracy into sequential, interdependent decisions. This process enables the model to implicitly understand causal patterns in urban flow. While the cold-start model captures static spatial relationships (e.g., POI connectivity), the RL phase learns adaptive policies for temporal dynamics, which may vary across cities. For instance, RL may uncover that San Francisco's ferry-dependent commute patterns require different temporal attention than Los Angeles' freeway-dominated traffic. Policy gradient updates selectively reinforce trajectory predictions that maintain spatiotemporal consistency over multiple steps, smoothing out erratic single-step predictions from cold-start initialization. 

However, the effectiveness of this approach depends on two critical factors: (1) The cold-start model must provide sufficiently accurate initial state representations for RL to explore meaningful action spaces, and (2) The reward function must balance immediate accuracy with long-term consistency, such as sparse rewards. Our experiments suggest that the current formulation achieves this balance, though future work could explore curriculum-based RL to progressively introduce complex temporal dependencies. These findings position RL\&CS as a principled framework for spatiotemporal transfer learning, where representation preservation and adaptive control mutually enhance cross-city generalization.

\subsection{Cross-Size and Cross-Type Adaptation: From Fresno to Los Angeles}

\begin{table}[t]
\centering
\caption*{\textbf{Table 3 \textbar{} Cross-Size and Cross-Type Adaptation: From Fresno to Los Angeles}}
\label{tab:fresno_to_la_comparison}
\resizebox{\textwidth}{!}{%
\begin{tabular}{
>{\centering\arraybackslash}m{2.4cm}  
*{1}{>{\centering\arraybackslash}m{2.6cm}} 
*{2}{>{\centering\arraybackslash}m{2.3cm}} 
*{1}{>{\centering\arraybackslash}m{2.6cm}} 
*{1}{>{\centering\arraybackslash}m{3.2cm}} 
}
\toprule
\textbf{Metric} & \textbf{RL\&CS} & \textbf{SL} & \textbf{CS} & \textbf{SL \& CS} & \textbf{FRE Model on LA} \\
\midrule
MSE & \textbf{1.86} & 1.79 & 1.87 & 1.94 & 3.07 \\
MAE & \textbf{1.70} & 2.08 & 1.79 & 2.08 & 1.43 \\
\bottomrule
\addlinespace[4pt]
\end{tabular}%
}
\vspace{4pt}
\noindent\footnotesize
\raggedright
\parbox{\textwidth}{
Note: LA = Los Angeles, FRE = Fresno. CS = Cold Start, SL = Supervised Learning. ``SL \& CS" denotes supervised learning fine-tuning with cold start initialization. ``RL \& CS" denotes reinforcement learning fine-tuning with cold start initialization.  MSE and MAE values are multiplied by $10^{-2}$ for readability. Bold values indicate best performance; underline indicates the second best.
}

\end{table}
The Fresno-to-Los Angeles case is deliberately chosen to test extreme scaling challenges, where the agricultural city's sparse mobility patterns must adapt to Los Angeles's complex metro system, probing the model's capacity for hierarchical feature transfer across urban tiers.

The cross-city comparison reveals distinct transfer learning patterns between the two scenarios. When transferring from Los Angeles to San Francisco, the pre-trained Los Angeles model shows reasonable zero-shot performance, indicating that large-city training captures broadly applicable urban mobility patterns. In contrast, transferring from Fresno to Los Angeles fails catastrophically, demonstrating that small-city models cannot generalize to metro-scale dynamics. This asymmetry stems from fundamental differences in urban complexity—Los Angeles's training data encompasses diverse spatiotemporal patterns that transfer down the urban hierarchy, while Fresno's limited scope cannot scale upward. 

The CS approach demonstrates adaptive effectiveness based on transfer direction. When transferring from Los Angeles to San Francisco, CS's selective fine-tuning of deep layers preserves transferable spatial features while adapting city-specific patterns. However, in Fresno-to-Los Angeles transfers, CS's conservative nature limits its ability to capture Los Angeles's complex dynamics, though it maintains stability. This creates a valuable trade-off: CS reliably predicts typical flows while being slightly conservative on outliers, often preferable for real-world deployment where consistent performance outweighs perfect outlier prediction. The method acts as an adaptive information bottleneck - ideal for ``downward" transfers (large-to-small) where it protects core spatial knowledge, but somewhat restrictive for ``upward" scaling. This reveals CS's strategic role: not a universal solution, but a versatile tool whose effectiveness depends on the urban hierarchy between source and target cities. Its true strength lies in balanced adaptation - maintaining robust performance across scenarios rather than optimizing for any single metric.

The RL\&CS framework demonstrates substantial improvements over supervised learning, achieving an 18.27\% reduction in MAE while maintaining comparable MSE. This enhancement arises from complementary mechanisms: CS adaptation provides a 13.94\% MAE reduction by preserving transferable spatial patterns through constrained fine-tuning, while reinforcement learning contributes an additional 5.03\% MAE improvement through temporal policy optimization. Our carefully designed reward function, combining flow-weighted MSE and MAE components, drives this temporal refinement by emphasizing high-flow edges (via the flow-sensitive weighting in the MSE term) and outlier robustness. The sparse, episode-wise reward structure guides RL to optimize for long-term consistency rather than point-wise accuracy, explaining the framework's particular effectiveness in reducing large prediction errors - a critical requirement for real-world deployment where avoiding severe flow underestimations is paramount. These results demonstrate how the hybrid approach successfully balances spatial knowledge preservation with temporal adaptation, with CS establishing robust baseline predictions that RL further refines through its focus on critical routes and temporal patterns.

\subsection{Ablation and Component Analysis}
We evaluate the impact of removing key architectural or training components from UrbanPulse. The ablation study results (Table 4) reveals that each component of UrbanPulse contributes meaningfully to its performance, with the full model consistently outperforming ablated versions across all evaluated cities. The removal of POI embeddings leads to the most pronounced degradation, particularly in smaller cities like Fresno, where the MAE nearly doubles. This suggests that fine-grained spatial representations are critical for capturing unique urban mobility patterns, especially in regions where aggregated data may obscure important local variations. Similarly, disabling temporal encoding results in a substantial increase in prediction error, particularly in Los Angeles, where complex, time-dependent flow dynamics require explicit modeling of periodic patterns such as rush hours or weekend activity shifts. Meteorological features also play a notable role, with their absence disproportionately affecting smaller cities, likely because weather conditions—such as rain or extreme heat—can more significantly alter mobility behavior in less densely populated areas compared to large metros with robust transportation networks.

The success of reinforcement learning fine-tuning further underscores the importance of adaptive training strategies. When transferring the model from Los Angeles to San Francisco, the RL-enhanced version achieves a measurable improvement in accuracy, demonstrating its ability to refine predictions by learning city-specific mobility nuances. Interestingly, while temporal convolutions are essential for handling Los Angeles’s intricate traffic dynamics, their impact is minimal in Fresno, implying that smaller cities may exhibit simpler temporal patterns that do not require deep hierarchical modeling. These findings collectively validate UrbanPulse’s design, which balances universal architectural strengths—such as its graph-based encoder-decoder structure—with flexible, context-aware components that can be emphasized or de-emphasized based on the target city’s characteristics. The results suggest that future urban prediction systems could benefit from dynamic feature weighting mechanisms to automatically adjust model focus depending on the urban environment.

\begin{table}[t]
\centering
\caption*{\textbf{Table 4 \textbar{} Ablation study results across three cities}}
\label{tab:ablation_all}
\resizebox{\textwidth}{!}{%
\begin{tabular}{
>{\centering\arraybackslash}m{2.4cm}  
>{\centering\arraybackslash}m{1.4cm}  
*{7}{>{\centering\arraybackslash}m{1.4cm}}  
}
\toprule
\textbf{Target City} & \textbf{Metric} & \textbf{Full} & \textbf{No PE} & \textbf{No MF} & \textbf{No TE} & \textbf{No TC} & \textbf{No RL} \\
\midrule
\multirow{2}{*}{\textbf{LA (Source)}} & MSE & \textbf{1.79} & 1.93 & 2.05 & 2.14 & 2.28 & -- \\
                                      & MAE & \textbf{2.08} & 2.76 & 2.24 & 3.28 & 2.64 & --\\
\midrule
\multirow{2}{*}{\textbf{SF}} & MSE & \textbf{0.64} & -- & -- & -- & -- & 0.68 \\
                             & MAE & \textbf{1.08} & -- & -- & -- & --  & 1.13 \\
\midrule
\multirow{2}{*}{\textbf{FRE}} & MSE & \textbf{0.39} & 0.44 & 0.41 & 0.40 & 0.39 & -- \\
                              & MAE & \textbf{0.73} & 1.97 & 1.39 & 1.23 & 0.95 & -- \\
\midrule
\multirow{2}{*}{\textbf{LA (Target)}} & MSE & \textbf{1.86} & -- & -- & -- & -- & 1.87 \\
                                      & MAE & \textbf{1.70} & -- & -- & -- & -- & 1.79 \\
\bottomrule
\addlinespace[4pt]
\end{tabular}%
}
\vspace{4pt}
\noindent\footnotesize
\raggedright
\parbox{\textwidth}{
Note: PE = POI Embedding, MF = Meteorological Features, TE = Temporal Encoding, TC = Temporal Convolution, CS = Cold Start, RL = Reinforcement Learning. MSE and MAE values are multiplied by $10^{-2}$ for readability. “--” indicates the setting is not applicable. LA (Source) refers to the pre-trained model, while LA (Target) indicates the fine-tuned model after transfer. Bold values indicate best performance for each city.
}
\end{table}

\section{Discussion}
UrbanPulse demonstrates that hierarchical spatiotemporal modeling, combined with adaptive transfer learning, effectively addresses key challenges in urban flow prediction. Its encoder–decoder architecture captures both local and global dynamics: temporal encoding and convolution layers extract short-term fluctuations, while spatiotemporal graph convolutions model mobility patterns conditioned on evolving network structures. A Transformer-based decoder then attends across space and time to capture long-range dependencies, enabling robust multi-scale forecasting. Despite strong improvements in MSE, the narrower MAE margin highlights ongoing difficulty in predicting rare or extreme flows under low-signal conditions.

The three-stage transfer framework proves particularly effective for cross-city adaptation. Results show that transferability is shaped less by data volume and more by the structural complexity of the source city. Simple cities with limited spatial heterogeneity and temporal variability fail to generalize upward to metropolitan systems, whereas models trained on complex cities adapt well to smaller ones. This asymmetry underscores the importance of urban hierarchy in transfer learning.

The CS mechanism provides a pragmatic middle ground: by selectively fine-tuning deep layers, it preserves transferable representations while enabling adaptation to target-specific features. Its stable performance across transfer directions makes it well suited for real-world applications that demand reliability over aggressive specialization.

RL further enhances this framework by introducing temporal policy optimization. Instead of minimizing a pointwise loss, RL uses a multi-objective reward function combining flow-weighted MSE and MAE, allowing the model to prioritize major flow routes while remaining robust to rare but critical deviations. Through temporal credit assignment and multi-step feedback, RL smooths predictions and mitigates cold-start limitations in dynamic settings. Together, CS and RL create a complementary framework—preserving structural priors while enabling adaptive temporal control—crucial for robust generalization across diverse urban systems.

UrbanPulse’s modular design supports scalable deployment by adjusting the depth of temporal and graph convolutions or transformer attention layers based on task complexity. However, two limitations remain: (1) the reward function may underweight low-flow but structurally important edges, and (2) the 5\% MAE improvement from RL over CS alone may not justify the additional computational cost in all settings.

Future work should emphasize uncertainty quantification for policy-relevant scenarios \cite{steentoft2024quantifying, huang2023uncertainty}. Lightweight strategies such as Bayesian last-layer inference \cite{watson2021latent} or quantile regression \cite{rodrigues2020beyond,yu2001bayesian} could yield predictive intervals without architectural changes. The edge-centric design also opens paths for anomaly detection by monitoring attention shifts during disruptions such as extreme weather or pandemics.

UrbanPulse’s broader significance lies in bridging AI and urban science. By modeling cities as learnable spatiotemporal graphs, it provides a testbed for hypothesis-driven experimentation—enabling simulations of traffic interventions, infrastructure shocks, or behavioral shifts with fine-grained interpretability. In doing so, UrbanPulse positions machine learning not just as a forecasting tool, but as a platform for discovery in the science of cities.

\section{Methodology}

\subsection{Spatiotemporal Graph Formulation}

To model the evolving dynamics of urban mobility, we represent the city as a sequence of dynamic graphs that capture both the spatial distribution of population and movement patterns over time. Each graph corresponds to a discrete time step, associated with a fixed-length temporal interval.

\paragraph{Data.}
We use anonymized GPS mobility traces from March 2023, collected by CITYDATA.ai across three metropolitan areas: Los Angeles, San Francisco, and Fresno.The raw dataset consists of 1.10 billion points, which is reduced to 103.07 million after cleaning. Additionally, we apply speed-based outlier removal during graph generation to ensure the graph's accuracy and reliability. These traces are processed into time-indexed sequences of population flow graphs, with each graph snapshot corresponding to a 15-minute interval. The real-world points of interest (POIs) are extracted from OpenStreetMap using the Python package OSMnx. We obtain meteorological data from the ERA5 reanalysis dataset provided by the Copernicus Climate Data Store, including air temperature at 2 meters above ground at 2 meters above ground, total precipitation, and 10-meter eastward and northward wind components. Wind speed is computed as the Euclidean norm of the two wind vectors. The Los Angeles dataset is used for pre-training, and the San Francisco dataset is reserved for transfer evaluation. For pre-training, the graph sequences are divided into training, validation, and test sets in approximate proportions of 70\%, 15\%, and 15\%. For the San Francisco dataset, the data used for cold-start fine-tuning, reinforcement learning-based adaptation, and final evaluation is split in a ratio of approximately 2:4:1.

The timeline is discretized into fixed-length intervals of duration \(\Delta t\), each represented as a time step \(t\). For each time step \(t\) (i.e., interval \([t, t+\Delta t)\)), we construct a directed graph snapshot \(\mathcal{G}_t\) that captures the population distribution and mobility patterns observed during that period. User locations are mapped to the nearest POI using spatial proximity, attributing each individual to a single node. This assignment yields per-POI population counts and enables the inference of transitions between POIs, which are used to define edge connections and flow magnitudes.

\paragraph{Node representation.}
Each graph snapshot consists of a fixed set of nodes representing real-world points of interest (POIs) distributed across the city. For each node, we construct a multi-dimensional feature vector that captures both static and dynamic attributes. These include a categorical encoding of POI type, geographic coordinates (latitude and longitude), the number of people assigned to the node at each time interval, and three meteorological variables: 2-meter air temperature, total precipitation, and wind speed. 

To obtain the population count at each POI, individual users are assigned to their nearest POI based on their first GPS observation within each 15-minute interval. This one-to-one assignment avoids duplication and yields consistent node-level population estimates. All real-value variables are normalized to the range [0,1] using min–max scaling based on training data statistics. To reduce the influence of extreme values, precipitation is first capped at 50 mm/h and then log-transformed before normalization. In addition to aligning heterogeneous feature scales and enabling efficient model training, normalization also facilitates transferability across cities with differing spatial layouts and population densities. In particular, normalizing geographic coordinates preserves the relative spatial configuration of POIs, allowing the model to reason about urban structure in a city-agnostic manner.

The inclusion of weather-related features is essential for capturing external environmental influences on human movement. Temperature and precipitation, in particular, strongly affect the temporal variability of mobility flows, while wind speed serves as a proxy for physical discomfort. Integrating these exogenous signals allows the model to better anticipate short-term fluctuations in population flow under varying conditions.

\paragraph{Edge construction.}
Directed edges \((v_i, v_j) \in \mathcal{E}_t\) capture observed transitions between POIs within each interval. An edge is created if at least one user moves from POI \(i\) to POI \(j\) at time step \(t\). The corresponding edge weight \(w_{ij}^t\) denotes the number of such transitions. These transitions are inferred by comparing consecutive user locations, where a move from one POI to another within the same user trajectory is counted as a valid flow.  Each edge is assigned a weight \(w_{ij}^t\), corresponding to the total number of users making that transition at time step \(t\). To ensure spatial plausibility, transitions that exceed a city-specific distance threshold or violate realistic travel speeds are filtered out.

\paragraph{Input preparation.}

To construct training examples, we apply a sliding window over the sequence of dynamic graphs. At each step, a fixed-length sub-sequence of consecutive snapshots is used as input, and the edge flows at the final time step serve as the prediction target.

Each training batch consists of \(B\) such windows, each spanning \(T\) time steps. For every window, node features from all graphs are aggregated into a tensor \(\mathbf{X} \in \mathbb{R}^{B \times N \times F \times T}\), where \(N\) denotes the number of POIs and \(F = 7\) represents the node feature dimension. Among these features, the POI type index serves as a categorical descriptor of each location's urban function. These indices are transformed into dense, trainable vectors via a POI embedding layer, which captures functional similarities and distinctions across POI categories. The resulting embeddings are concatenated with other normalized node-level features and passed through a temporal encoding module to form a unified input representation.

This representation captures the spatiotemporal dynamics of activity at each POI and serves as the input to the temporal graph encoder. By leveraging overlapping windows, the model is exposed to diverse temporal contexts, thereby improving data efficiency and enabling high-resolution forecasting of urban population flows.

\paragraph{Prediction objective.}
The core task is to predict edge-level population flows between all POI pairs over time. Formally, we aim to learn a function that maps a sequence of recent graphs to the future flow values:
\begin{equation}
    f_\Theta: \left\{ \mathcal{G}_{t-k}, \ldots, \mathcal{G}_{t} \right\} \mapsto \hat{\mathcal{W}} = \{ \hat{w}_{ij}^t \mid (i, j) \in \mathcal{V} \times \mathcal{V},\ t = 1, \ldots, T \},
\end{equation}
where \(\Theta\) denotes model parameters, \(\hat{w}_{ij}^t\) is the predicted number of individuals moving from POI \(i\) to POI \(j\) during interval \(t\), and \(\hat{\mathcal{W}}\) is the full set of predicted edge flows across all time steps. This formulation enables fine-grained forecasting by capturing both spatial dependencies and temporal dynamics in urban environments.

\subsubsection{Temporal Graph Encoder}

The encoder processes node-level dynamics through four sequential components:

\paragraph{(1) Temporal Encoding.}  
To encode the temporal position of each node feature, we incorporate fixed sinusoidal positional encodings \(\mathbf{PE}(t)\). 
Node feature at time $t$ is augmented with sinusoidal positional encodings to capture temporal order:
\begin{equation}
    \mathbf{Z} = \mathbf{X} + \mathbf{PE}(t),
\end{equation}
This technique helps the model distinguish among different positions in the input sequence and has proven effective in sequence modeling tasks. 

\paragraph{(2) Temporal Convolution.}
To capture short-range temporal dependencies, we apply a stack of one-dimensional convolutional layers along the time axis of the node features. Each layer enables a node to integrate information from its recent history by aggregating signals across neighboring time steps. At each layer \(l\), the temporal features are updated as:
\begin{equation}
\mathbf{H}^{(l)} = \mathrm{Dropout} \left( \mathrm{LayerNorm} \left( \mathrm{ReLU} \left( \mathbf{H}^{(l-1)} \ast \mathbf{W}_{e}^{(l)} \right) \right) \right), \quad l = 1, \dots, L_e,
\end{equation}
where the input \(\mathbf{H}^{(0)} = \mathbf{Z}\) and \(\mathbf{W}_{e}^{(l)}\) denotes the learnable convolution weight. The ReLU activation introduces non-linearity, allowing the model to learn complex temporal patterns, while layer normalization stabilizes training by normalizing feature responses. Dropout is applied for regularization, helping prevent overfitting during training. The module produces temporally enriched node features \(\mathbf{H}^{(L_e)}\).

\paragraph{(3) Spatio-temporal Graph Convolution.}
To capture spatial dependencies, we apply graph convolutional layers independently at each time step. Each graph convolution integrates neighborhood information and combines it with a residual connection to stabilize training and retain useful features from previous layers. At each time step \(t\) and for each batch \(b\), the node embeddings are updated as:
\begin{equation}
    \mathbf{H}_{b,t}^{(l)} = \mathrm{ReLU}(\hat{\mathbf{A}}_{b,t} \mathbf{H}_{b,t}^{(l-1)} \mathbf{W}_g^{(l)}) + \mathbf{R}^{(l)} \mathbf{H}_{b,t}^{(l-1)}, \quad l = 1, \dots, L_g,
\end{equation}
where \(\mathbf{H}_{b,t}^{(0)} = \mathbf{H}^{(L_e)}[b, :, :, t]\), \(\hat{\mathbf{A}}_{b,t}\) is the normalized adjacency matrix, \(\mathbf{W}_g^{(l)}\) are learnable graph convolution weights, and \(\mathbf{R}^{(l)}\) projects the residual connection. The final output is the refined node representations \(\mathbf{H}_{b,t}^{(L_g)}\) that encode spatial structure conditioned on recent temporal context.

\paragraph{(4) Edge Feature Generation.}
To enable edge-wise flow prediction, we construct directed edge embeddings from the final node representations. For each edge \((i, j) \in \mathcal{E}_{b,t}\), we retrieve the corresponding node embeddings \(\mathbf{h}_i^{b,t}\) and \(\mathbf{h}_j^{b,t}\) from the output of the final graph convolutional layer. A composite edge feature is formed by concatenating the source and target embeddings, along with their element-wise absolute difference:
\begin{equation}
\mathbf{u}_{ij}^{b,t} = [\mathbf{h}_i^{b,t} \| \mathbf{h}_j^{b,t} \| |\mathbf{h}_i^{b,t} - \mathbf{h}_j^{b,t}|].
\end{equation}
This feature vector is passed through a shared linear projection to obtain the edge embedding \(\mathbf{e}_{ij}^{b,t} \in \mathbb{R}^{D_g}\), where \(D_g\) is the embedding dimension. The resulting edge embeddings are stored in a tensor and subsequently flattened into \(\mathbf{E}\), which is organized as a sequence of tokens spanning time steps and batches. A binary mask \(\mathbf{A}\) indicates valid edge entries across time steps and is used during attention to prevent attending to padded positions. Together, \(\mathbf{E}\) and \(\mathbf{A}\) serve as inputs to the Transformer-based decoder.

\subsubsection{Transformer-Based Decoder}

The decoder models temporal and pairwise interactions between POIs to infer future population flows. It processes the edge embedding sequence using a stack of Transformer layers, with the binary mask \(\mathbf{A}\) applied during attention to ignore padded positions. By operating over all edges across time and space, the decoder captures dynamic dependencies in mobility patterns.

\paragraph{Transformer Processing.} 
The edge embedding sequence is processed by a Transformer decoder composed of \(L_d\) layers. Each layer consists of a multi-head self-attention (MHA) mechanism followed by a position-wise feedforward network (FFN), with both submodules wrapped in residual connections and followed by layer normalization. The MHA block models temporal and pairwise interactions between edge embeddings, while the FFN enhances representational capacity through two linear projections with a ReLU activation in between. Dropout is applied after each submodule for regularization. The decoder input is initialized as \(\mathbf{Z}^{(0)} = \mathbf{E}\) with binary mask  \(\mathbf{A}\), and the final output \(\mathbf{Z}^{(L_d)}\) preserves the dimensionality of the edge embedding space.

\paragraph{Flow Prediction.}
The output \(\mathbf{Z}^{(L_d)}\) is reshaped to \(\mathbb{R}^{T \times B \times M \times D_g}\), and flow values are predicted for each valid edge using a shared linear transformation followed by a Softplus activation:

\begin{equation}
\hat{\mathcal{W}} = \mathrm{Softplus} \left( \mathbf{W}_{\mathrm{out}} \mathbf{z}^{(L_d)} + \mathbf{b}_{\mathrm{out}} \right),
\end{equation}
where \(\mathbf{W}_{\mathrm{out}}\) and \(\mathbf{b}_{\mathrm{out}}\) are the learnable weights and bias of the final output layer. The \(\mathrm{Softplus}\) activation ensures non-negative predictions, and the output corresponds to the magnitude of population transfer. Directionality is captured by the edge indices, which define the directed graph and are reflected in the asymmetric normalized adjacency matrix.

\subsection{Three-Stage Transfer Learning}

To enable robust cross-city generalization and adaptive fine-tuning, we design a three-stage transfer learning procedure: (1) supervised pre-training on a source city, (2) cold-start fine-tuning on a data-scarce target city, and (3) reinforcement learning-based refinement through reward-driven adaptation. 

\subsubsection{Stage 1: Pre-training.}
In the first stage, the model is pre-trained on large-scale population flow data from the source city—Los Angeles—to learn transferable spatiotemporal representations that generalize across urban environments. The encoder and decoder are jointly optimized in an end-to-end manner to minimize the mean squared error between predicted and observed inter-POI flows. To improve generalization and mitigate overfitting, we apply weight decay in conjunction with standard \(L_2\) regularization across all trainable parameters. Training is performed using an adaptive gradient-based optimizer with a small learning rate to ensure stable convergence. To further enhance robustness, gradient clipping is employed to prevent exploding gradients during early training epochs. All layers of the model are fully trainable in this stage, allowing the encoder to extract latent spatiotemporal features and the Transformer-based decoder to learn flow prediction dynamics at scale.

\subsubsection{Stage 2: Cold-start fine-tuning.}
In the second stage, the pre-trained model is adapted to a data-scarce target city—San Francisco—through cold-start fine-tuning. This setting simulates realistic deployment conditions where only limited labeled flow data is available in the target domain. Rather than reinitializing the model or retraining all parameters, we selectively fine-tune a subset of the architecture while freezing the remaining layers. Concretely, we freeze the early layers of the temporal convolution module, the graph convolutional module, and the Transformer decoder. These components are designed to capture generalizable temporal patterns and topological structures that are expected to transfer across cities. Fine-tuning is applied only to the upper layers of the temporal and graph convolution modules, the Transformer decoder, and the final output projection. This selective adaptation enables the model to specialize to local mobility dynamics while preserving shared spatiotemporal priors and mitigating overfitting to limited target data.

\subsubsection{Stage 3: Reinforcement learning-based fine-tuning.}

We formulate the task of model adaptation for urban population flow prediction as a sequential Markov decision process (MDP), where a reinforcement learning agent interacts with a structured simulation environment to optimize model behavior through iterative feedback. At each step, the agent observes a state—a compact, structured representation of the model’s current predictive performance and spatiotemporal graph characteristics. Based on this observation, the agent selects actions that adaptively modifies internal parameters of the output layer of the decoder responsible for estimating edge-level population flows. Following each action, the agent receives a mixed reward that reflects the accuracy of the model’s predictions, emphasizing both overall error and performance on high-flow, high-impact edges. The environment transitions by updating the model’s internal state and advancing to the next input sequence, allowing the agent to continuously refine its strategy over time. 

\paragraph{State.}
The state is represented as a 1,560-dimensional vector, which encodes both the dynamic evolution and structural characteristics of the urban mobility graph. This vector aggregates statistical features extracted from the intermediate representations of the spatiotemporal graph neural network, ensuring a compact and informative encoding of the system’s current state. Temporal dynamics are captured by computing the mean, standard deviation, maximum, and minimum of edge embedding differences across consecutive time steps, reflecting the evolution of traffic patterns. These temporal statistics provide a 1,024-dimensional representation that captures short-term flow variations across the entire graph. To incorporate the urban infrastructure context, node-level features are represented by their global mean and standard deviation, summarizing the distribution of node attributes. Additionally, edge embeddings are characterized by their mean and standard deviation, contributing a 512-dimensional snapshot of the graph’s relational structure at each time step. We set the data loader stride to a fraction of the sliding window size during RL fine-tuning to balance data continuity and diversity while mitigating state overfitting. The final state vector is standardized for stable training, integrating these components into a permutation-invariant representation. This approach ensures the model remains agnostic to graph size, while preserving essential spatiotemporal dependencies critical for mobility forecasting. 

\paragraph{Action.}
At each decision step, the reinforcement learning agent produces a continuous action vector that directly adjusts the parameters of the model’s final output layer. Rather than generating discrete decisions, the agent outputs a real-valued vector that acts as a set of scaling factors, modulating groups of weights within the decoder's output projection. The scaling action mode outperforms additive updates by preserving pretrained feature relationships through proportional weight adjustments. Its multiplicative nature ensures model stability while allowing controlled fine-tuning, particularly for sensitive parameters. The constrained update range prevents disruptive parameter shifts, while still supporting effective policy learning. In this paper, the output layer comprises 256 weights and a separate bias term. The weights are divided into 8 or 16 contiguous blocks, each corresponding to a distinct action, with an additional action assigned to the bias term. Each element of the action vector corresponds to a specific block, enabling structured, group-wise modifications, with parameters clamped within the range of -2 to 2. This design allows for expressive control over the model's predictive behavior while maintaining sample efficiency, avoiding the instability that may arise from modifying each weight independently. By adapting the output layer in this targeted way, the agent can fine-tune the model’s predictions in response to changing patterns in the data. The completed fine-tuning process can be formulated as follows:
\begin{equation}
    \mathbf{W}_{\mathrm{out}}^{(t+1)} = \mathbf{W}_{\mathrm{out}}^{(t)} \odot \left(1 + \mathbf{G} \mathbf{a}_t \right),
\end{equation}
where $\mathbf{W}_{\mathrm{out}}^{(t)}$ denotes the decoder's output layer weights at step $t$, $\mathbf{a}_t \in \mathbb{R}^{33}$ is the continuous action vector generated by the agent, and $\mathbf{G} \in \mathbb{R}^{256 \times 33}$ is a fixed binary grouping matrix that maps each action dimension to a contiguous block of output weights. The element-wise multiplication $\odot$ enables block-wise scaling, allowing the agent to adjust decoder behavior in a structured yet efficient manner.

\paragraph{Reward.}  
To guide the agent toward high-quality predictions, we design a sparse reward function that provides feedback only at the end of each episode. The reward penalizes both the magnitude and frequency of prediction errors, placing greater emphasis on high-flow edges. Specifically, it is formulated as a weighted combination of mean squared error and mean absolute error (MAE), where the MSE is scaled by a flow-dependent weight. This weighting increases the penalty for inaccuracies on edges with higher true values, encouraging the agent to focus on areas with dense population movement. The MAE component adds robustness to outliers and complements the sensitivity of the MSE. This reward design aligns the reinforcement learning objective with the real-world need to accurately capture rare but impactful mobility patterns. By providing feedback only at the end of each episode, the sparse scheme encourages the agent to develop long-term strategies while maintaining practical relevance. The reward function is defined as:
\begin{align}
    r_t &= -\delta (e_{\text{wmse}, t} + \lambda e_{\mathrm{mae}, t}), \\
    e_{\text{wmse}, t} &= \frac{1}{N_b} \sum_{i=1}^{N_b} \left(1 + \frac{w_{ij}^t}{w_{\mathrm{max}}} \right)(\hat{w}_{ij}^t - w_{ij}^t)^2,
\end{align}
where $\hat{w}_{ij}^t$ and $w_{ij}^t$ are the predicted and true edge values respectively for edge $(i,j)$ at time $t$, $e_{\mathrm{wmse}, t}$ is the weighted mean squared error defined as (11), and $e_{\mathrm{mae}, t}$ is the MAE error. $N_b$ denotes the number of edges in the current batch. $w_{\mathrm{max}}$ is the maximum ground-truth edge value in the batch, used to normalize flow magnitudes. The weighting term $\left(1 + \frac{w_{ij}^t}{w_{\mathrm{max}}} \right)$ increases the contribution of high-flow edges to the error. $\delta$ is a global scaling factor, and $\lambda$ controls the relative importance of MAE in the overall reward.

\paragraph{Learning algorithm.}

To enable stable and sample-efficient policy learning in a high-dimensional and continuous action space, we adopt the Proximal Policy Optimization (PPO) algorithm \cite{schulman2017proximal} as our reinforcement learning backbone. PPO optimizes a stochastic policy by maximizing a clipped objective function, which ensures that updates to the policy are not too large and that training remains stable. The actor is updated to maximize the expected returns while maintaining the policy's proximity to the previous policy, according to the loss:
\begin{equation}
\mathcal{L}_{\mathrm{actor}} = \mathbb{E}_{\mathbf{s}_t, \mathbf{a}_t} \left[ \min \left( \rho_t \cdot \hat{A}_t, \text{clip}(\rho_t, 1 - \epsilon, 1 + \epsilon) \cdot \hat{A}_t \right) - c_1 \left(V(\mathbf{s}_t) - V_{\text{target}(\mathbf{s}_t)} \right)^2 + c_2 \cdot H \right]
\end{equation}
where $\rho_t$ is the probability ratio between the new and old policies, $\hat{A}_t$ is the advantage estimate at time step $t$, and $\epsilon$ is a hyperparameter that controls the clipping range. The clipping operation ensures that large updates that could destabilize training are avoided, promoting more conservative updates and improving sample efficiency. $c_1$ and $c_2$ are the coefficients for the value function and entropy, respectively. $H$ represents the policy entropy.

The value function $V(\mathbf{s}_t)$ is trained using the mean squared error between the predicted value and the target value, which is computed as:
\begin{equation}
\mathcal{L}_{\mathrm{critic}} = \mathbb{E}_{\mathbf{s}_t} \left[ \left( V(\mathbf{s}_t) - \hat{R}_t \right)^2 \right],
\end{equation}
where $\hat{R}_t$ is the reward-to-go at time step $t$, which serves as the target value for the critic.

\bibliographystyle{unsrtnat}
\bibliography{references} 

\begin{thebibliography}{51}
\providecommand{\natexlab}[1]{#1}
\providecommand{\url}[1]{\texttt{#1}}
\expandafter\ifx\csname urlstyle\endcsname\relax
  \providecommand{\doi}[1]{doi: #1}\else
  \providecommand{\doi}{doi: \begingroup \urlstyle{rm}\Url}\fi

\bibitem[Rong et~al.(2024)Rong, Ding, and Li]{rong2024interdisciplinary}
Can Rong, Jingtao Ding, and Yong Li.
\newblock An interdisciplinary survey on origin-destination flows modeling:
  Theory and techniques.
\newblock \emph{ACM Computing Surveys}, 57\penalty0 (1):\penalty0 1--49, 2024.

\bibitem[Sayarshad(2024)]{sayarshad2024optimization}
Hamid~R. Sayarshad.
\newblock Optimization of electric charging infrastructure: integrated model
  for routing and charging coordination with power-aware operations.
\newblock \emph{npj Sustainable Mobility and Transport}, 1:\penalty0 4, 2024.
\newblock \doi{10.1038/s44333-024-00004-6}.

\bibitem[Tang et~al.(2024)Tang, Mao, Liu, Liu, Wang, and Huang]{tang2024origin}
Tianli Tang, Jiannan Mao, Ronghui Liu, Zhiyuan Liu, Yiran Wang, and Di~Huang.
\newblock Origin-destination matrix prediction in public transport networks:
  Incorporating heterogeneous direct and transfer trips.
\newblock \emph{IEEE Transactions on Intelligent Transportation Systems}, 2024.

\bibitem[Cabanas-Tirapu et~al.(2025)Cabanas-Tirapu, Dan{\'u}s, Moro,
  Sales-Pardo, and Guimer{\`a}]{cabanas2025human}
Oriol Cabanas-Tirapu, Llu{\'\i}s Dan{\'u}s, Esteban Moro, Marta Sales-Pardo,
  and Roger Guimer{\`a}.
\newblock Human mobility is well described by closed-form gravity-like models
  learned automatically from data.
\newblock \emph{Nature Communications}, 16\penalty0 (1):\penalty0 1336, 2025.

\bibitem[Guo et~al.(2025)Guo, Tan, Tang, and Shi]{guo2025unifying}
Qi~Guo, Qi~Tan, Jun Tang, and Benyun Shi.
\newblock Unifying spatiotemporal and frequential attention for traffic
  prediction.
\newblock \emph{Scientific Reports}, 15\penalty0 (1):\penalty0 953, 2025.

\bibitem[Gallotti et~al.(2024)Gallotti, Maniscalco, Barthelemy, and
  De~Domenico]{gallotti2024distorted}
Riccardo Gallotti, Davide Maniscalco, Marc Barthelemy, and Manlio De~Domenico.
\newblock Distorted insights from human mobility data.
\newblock \emph{Communications Physics}, 7\penalty0 (1):\penalty0 421, 2024.

\bibitem[Tan et~al.(2025)Tan, Huang, Batty, Li, Wang, Zhou, and
  Gong]{tan2025spatiotemporal}
Xingye Tan, Bo~Huang, Michael Batty, Weiyu Li, Qi~Ryan Wang, Yulun Zhou, and
  Peng Gong.
\newblock The spatiotemporal scaling laws of urban population dynamics.
\newblock \emph{Nature Communications}, 16\penalty0 (1):\penalty0 2881, 2025.

\bibitem[Rong et~al.(2023)Rong, Feng, and Ding]{rong2023goddag}
Can Rong, Jie Feng, and Jingtao Ding.
\newblock Goddag: Generating origin-destination flow for new cities via domain
  adversarial training.
\newblock \emph{IEEE Transactions on Knowledge and Data Engineering},
  35\penalty0 (10):\penalty0 10048--10057, 2023.

\bibitem[Shao et~al.(2025)Shao, Li, Zhang, Wang, and Li]{shao2025cross}
Junqi Shao, Shen Li, Ke~Zhang, Anyou Wang, and Meng Li.
\newblock Cross-city traffic prediction based on deep domain adaptive transfer
  learning.
\newblock \emph{Transportation Research Part C: Emerging Technologies},
  176:\penalty0 105152, 2025.

\bibitem[Zipf(1946)]{zipf1946p}
George~Kingsley Zipf.
\newblock The p 1 p 2/d hypothesis: on the intercity movement of persons.
\newblock \emph{American sociological review}, 11\penalty0 (6):\penalty0
  677--686, 1946.

\bibitem[Simini et~al.(2012)Simini, Gonz{\'a}lez, Maritan, and
  Barab{\'a}si]{simini2012universal}
Filippo Simini, Marta~C Gonz{\'a}lez, Amos Maritan, and Albert-L{\'a}szl{\'o}
  Barab{\'a}si.
\newblock A universal model for mobility and migration patterns.
\newblock \emph{Nature}, 484\penalty0 (7392):\penalty0 96--100, 2012.

\bibitem[Alessandretti et~al.(2020)Alessandretti, Aslak, and
  Lehmann]{Alessandretti2020}
Laura Alessandretti, Ulf Aslak, and Sune Lehmann.
\newblock The scales of human mobility.
\newblock \emph{Nature}, 587\penalty0 (7834):\penalty0 402--407, 2020.
\newblock \doi{10.1038/s41586-020-2909-1}.

\bibitem[Barbosa et~al.(2018)Barbosa, Barthelemy, Ghoshal, James, Lenormand,
  Louail, Menezes, Ramasco, Simini, and Tomasini]{Barbosa2018}
Hugo Barbosa, Marc Barthelemy, Gourab Ghoshal, Charlotte~R James, Maxime
  Lenormand, Thomas Louail, Ronaldo Menezes, Jos{\'e}~J Ramasco, Filippo
  Simini, and Marcello Tomasini.
\newblock Human mobility: Models and applications.
\newblock \emph{Physics Reports}, 734:\penalty0 1--74, 2018.

\bibitem[Schl{\"a}pfer et~al.(2021)Schl{\"a}pfer, Dong, O’Keeffe, Santi,
  Szell, Salat, Anklesaria, Vazifeh, Ratti, and West]{Schlapfer2021}
Markus Schl{\"a}pfer, Lei Dong, Kevin O’Keeffe, Paolo Santi, Michael Szell,
  Hadrien Salat, Samuel Anklesaria, Mohammad Vazifeh, Carlo Ratti, and
  Geoffrey~B West.
\newblock The universal visitation law of human mobility.
\newblock \emph{Nature}, 593\penalty0 (7860):\penalty0 522--527, 2021.

\bibitem[Song et~al.(2010)Song, Koren, Wang, and Barab{\'a}si]{Song2010}
Chaoming Song, Tal Koren, Pu~Wang, and Albert-L{\'a}szl{\'o} Barab{\'a}si.
\newblock Modelling the scaling properties of human mobility.
\newblock \emph{Nature physics}, 6\penalty0 (10):\penalty0 818--823, 2010.

\bibitem[Luca et~al.(2021)Luca, Barlacchi, Lepri, and
  Pappalardo]{luca2021survey}
Massimiliano Luca, Gianni Barlacchi, Bruno Lepri, and Luca Pappalardo.
\newblock A survey on deep learning for human mobility.
\newblock \emph{ACM Computing Surveys (CSUR)}, 55\penalty0 (1):\penalty0 1--44,
  2021.

\bibitem[Ke et~al.(2018)Ke, Yang, Zheng, Chen, Jia, Gong, and
  Ye]{ke2018hexagon}
Jintao Ke, Hai Yang, Hongyu Zheng, Xiqun Chen, Yitian Jia, Pinghua Gong, and
  Jieping Ye.
\newblock Hexagon-based convolutional neural network for supply-demand
  forecasting of ride-sourcing services.
\newblock \emph{IEEE Transactions on Intelligent Transportation Systems},
  20\penalty0 (11):\penalty0 4160--4173, 2018.

\bibitem[Zhang et~al.(2017)Zhang, Zheng, and Qi]{zhang2017deep}
Junbo Zhang, Yu~Zheng, and Dekang Qi.
\newblock Deep spatio-temporal residual networks for citywide crowd flows
  prediction.
\newblock In \emph{Proceedings of the AAAI conference on artificial
  intelligence}, volume~31, 2017.

\bibitem[Zhang et~al.(2019)Zhang, Zheng, Sun, and Qi]{zhang2019flow}
Junbo Zhang, Yu~Zheng, Junkai Sun, and Dekang Qi.
\newblock Flow prediction in spatio-temporal networks based on multitask deep
  learning.
\newblock \emph{IEEE Transactions on Knowledge and Data Engineering},
  32\penalty0 (3):\penalty0 468--478, 2019.

\bibitem[Lin et~al.(2019)Lin, Feng, Lu, Li, and Jin]{lin2019deepstn}
Ziqian Lin, Jie Feng, Ziyang Lu, Yong Li, and Depeng Jin.
\newblock Deepstn+: Context-aware spatial-temporal neural network for crowd
  flow prediction in metropolis.
\newblock In \emph{Proceedings of the AAAI conference on artificial
  intelligence}, volume~33, pages 1020--1027, 2019.

\bibitem[Hochreiter and Schmidhuber(1997)]{hochreiter1997long}
Sepp Hochreiter and J{\"u}rgen Schmidhuber.
\newblock Long short-term memory.
\newblock \emph{Neural computation}, 9\penalty0 (8):\penalty0 1735--1780, 1997.

\bibitem[Elman(1990)]{elman1990finding}
Jeffrey~L Elman.
\newblock Finding structure in time.
\newblock \emph{Cognitive science}, 14\penalty0 (2):\penalty0 179--211, 1990.

\bibitem[Ye et~al.(2021)Ye, Sun, Du, Fu, and Xiong]{ye2021coupled}
Junchen Ye, Leilei Sun, Bowen Du, Yanjie Fu, and Hui Xiong.
\newblock Coupled layer-wise graph convolution for transportation demand
  prediction.
\newblock In \emph{Proceedings of the AAAI conference on artificial
  intelligence}, volume~35, pages 4617--4625, 2021.

\bibitem[Liu et~al.(2023)Liu, Chen, Huang, Li, and Min]{liu2023gnn}
Jinbo Liu, Yunliang Chen, Xiaohui Huang, Jianxin Li, and Geyong Min.
\newblock Gnn-based long and short term preference modeling for next-location
  prediction.
\newblock \emph{Information Sciences}, 629:\penalty0 1--14, 2023.

\bibitem[Xue et~al.(2022)Xue, Jiang, Liang, Pang, Yabe, Ukkusuri, and
  Ma]{xue2022quantifying}
Jiawei Xue, Nan Jiang, Senwei Liang, Qiyuan Pang, Takahiro Yabe, Satish~V
  Ukkusuri, and Jianzhu Ma.
\newblock Quantifying the spatial homogeneity of urban road networks via graph
  neural networks.
\newblock \emph{Nature Machine Intelligence}, 4\penalty0 (3):\penalty0
  246--257, 2022.

\bibitem[Lee and Rhee(2022)]{lee2022ddp}
Kyungeun Lee and Wonjong Rhee.
\newblock Ddp-gcn: Multi-graph convolutional network for spatiotemporal traffic
  forecasting.
\newblock \emph{Transportation Research Part C: Emerging Technologies},
  134:\penalty0 103466, 2022.

\bibitem[Pareja et~al.(2020)Pareja, Domeniconi, Chen, Ma, Suzumura, Kanezashi,
  Kaler, Schardl, and Leiserson]{pareja2020evolvegcn}
Aldo Pareja, Giacomo Domeniconi, Jie Chen, Tengfei Ma, Toyotaro Suzumura,
  Hiroki Kanezashi, Tim Kaler, Tao Schardl, and Charles Leiserson.
\newblock Evolvegcn: Evolving graph convolutional networks for dynamic graphs.
\newblock In \emph{Proceedings of the AAAI conference on artificial
  intelligence}, volume~34, pages 5363--5370, 2020.

\bibitem[Zhao et~al.(2019)Zhao, Song, Zhang, Liu, Wang, Lin, Deng, and
  Li]{zhao2019t}
Ling Zhao, Yujiao Song, Chao Zhang, Yu~Liu, Pu~Wang, Tao Lin, Min Deng, and
  Haifeng Li.
\newblock T-gcn: A temporal graph convolutional network for traffic prediction.
\newblock \emph{IEEE transactions on intelligent transportation systems},
  21\penalty0 (9):\penalty0 3848--3858, 2019.

\bibitem[Wang et~al.(2021{\natexlab{a}})Wang, Lin, Guo, and Wan]{wang2021gsnet}
Beibei Wang, Youfang Lin, Shengnan Guo, and Huaiyu Wan.
\newblock Gsnet: Learning spatial-temporal correlations from geographical and
  semantic aspects for traffic accident risk forecasting.
\newblock In \emph{Proceedings of the AAAI conference on artificial
  intelligence}, volume~35, pages 4402--4409, 2021{\natexlab{a}}.

\bibitem[Jin et~al.(2023)Jin, Liu, Li, and Huang]{jin2023spatio}
Guangyin Jin, Lingbo Liu, Fuxian Li, and Jincai Huang.
\newblock Spatio-temporal graph neural point process for traffic congestion
  event prediction.
\newblock In \emph{Proceedings of the AAAI conference on artificial
  intelligence}, volume~37, pages 14268--14276, 2023.

\bibitem[Wang et~al.(2024)Wang, Zheng, Liu, Feng, Chen, Hao, and
  Song]{wang2024spatiotemporal}
Yu~Wang, Tongya Zheng, Shunyu Liu, Zunlei Feng, Kaixuan Chen, Yunzhi Hao, and
  Mingli Song.
\newblock Spatiotemporal-augmented graph neural networks for human mobility
  simulation.
\newblock \emph{IEEE Transactions on Knowledge and Data Engineering},
  36\penalty0 (11):\penalty0 7074--7086, 2024.

\bibitem[Cini et~al.(2023)Cini, Marisca, Bianchi, and Alippi]{cini2023scalable}
Andrea Cini, Ivan Marisca, Filippo~Maria Bianchi, and Cesare Alippi.
\newblock Scalable spatiotemporal graph neural networks.
\newblock In \emph{Proceedings of the AAAI conference on artificial
  intelligence}, volume~37, pages 7218--7226, 2023.

\bibitem[Wang et~al.(2019)Wang, Geng, Ma, Liu, and Yang]{wang2019cross}
Leye Wang, Xu~Geng, Xiaojuan Ma, Feng Liu, and Qiang Yang.
\newblock Cross-city transfer learning for deep spatio-temporal prediction.
\newblock In \emph{Proceedings of the 28th International Joint Conference on
  Artificial Intelligence}, pages 1893--1899, 2019.

\bibitem[Jin et~al.(2022)Jin, Chen, and Yang]{jin2022selective}
Yilun Jin, Kai Chen, and Qiang Yang.
\newblock Selective cross-city transfer learning for traffic prediction via
  source city region re-weighting.
\newblock In \emph{Proceedings of the 28th ACM SIGKDD conference on knowledge
  discovery and data mining}, pages 731--741, 2022.

\bibitem[Li et~al.(2024{\natexlab{a}})Li, Xia, Tang, Xu, Shi, Xia, Yin, and
  Huang]{li2024urbangpt}
Zhonghang Li, Lianghao Xia, Jiabin Tang, Yong Xu, Lei Shi, Long Xia, Dawei Yin,
  and Chao Huang.
\newblock Urbangpt: Spatio-temporal large language models.
\newblock In \emph{Proceedings of the 30th ACM SIGKDD Conference on Knowledge
  Discovery and Data Mining}, pages 5351--5362, 2024{\natexlab{a}}.

\bibitem[Liu et~al.(2024)Liu, Yang, Xu, Li, Long, Li, and Zhao]{liu2024spatial}
Chenxi Liu, Sun Yang, Qianxiong Xu, Zhishuai Li, Cheng Long, Ziyue Li, and Rui
  Zhao.
\newblock Spatial-temporal large language model for traffic prediction.
\newblock In \emph{2024 25th IEEE International Conference on Mobile Data
  Management (MDM)}, pages 31--40. IEEE, 2024.

\bibitem[Liang et~al.(2024)Liang, Liu, Wang, and Zhao]{liang2024exploring}
Yuebing Liang, Yichao Liu, Xiaohan Wang, and Zhan Zhao.
\newblock Exploring large language models for human mobility prediction under
  public events.
\newblock \emph{Computers, Environment and Urban Systems}, 112:\penalty0
  102153, 2024.

\bibitem[Li et~al.(2024{\natexlab{b}})Li, Azfar, and Ke]{li2024chatsumo}
Shuyang Li, Talha Azfar, and Ruimin Ke.
\newblock Chatsumo: Large language model for automating traffic scenario
  generation in simulation of urban mobility.
\newblock \emph{IEEE Transactions on Intelligent Vehicles}, 2024{\natexlab{b}}.

\bibitem[Tan et~al.(2024)Tan, Merrill, Gupta, Althoff, and
  Hartvigsen]{tan2024language}
Mingtian Tan, Mike~A Merrill, Vinayak Gupta, Tim Althoff, and Thomas
  Hartvigsen.
\newblock Are language models actually useful for time series forecasting?
\newblock In \emph{Proceedings of the 38th International Conference on Neural
  Information Processing Systems}, pages 60162--60191, 2024.

\bibitem[Manvi et~al.(2024)Manvi, Khanna, Mai, Burke, Lobell, and
  Ermon]{manvi2024geollm}
Rohin Manvi, Samar Khanna, Gengchen Mai, Marshall Burke, David~B Lobell, and
  Stefano Ermon.
\newblock Geollm: Extracting geospatial knowledge from large language models.
\newblock In \emph{ICLR}, 2024.

\bibitem[Zeng et~al.(2023)Zeng, Chen, Zhang, and Xu]{zeng2023transformers}
Ailing Zeng, Muxi Chen, Lei Zhang, and Qiang Xu.
\newblock Are transformers effective for time series forecasting?
\newblock In \emph{Proceedings of the AAAI conference on artificial
  intelligence}, volume~37, pages 11121--11128, 2023.

\bibitem[Chang et~al.(2024)Chang, Park, Ye, Yang, Seo, Chang, and
  Seo]{chang2024large}
Hoyeon Chang, Jinho Park, Seonghyeon Ye, Sohee Yang, Youngkyung Seo, Du-Seong
  Chang, and Minjoon Seo.
\newblock How do large language models acquire factual knowledge during
  pretraining?
\newblock \emph{Advances in neural information processing systems},
  37:\penalty0 60626--60668, 2024.

\bibitem[Wu et~al.(2019)Wu, Pan, Long, Jiang, and Zhang]{wu2019graph}
Zonghan Wu, Shirui Pan, Guodong Long, Jing Jiang, and Chengqi Zhang.
\newblock Graph wavenet for deep spatial-temporal graph modeling.
\newblock In \emph{Proceedings of the 28th International Joint Conference on
  Artificial Intelligence}, pages 1907--1913, 2019.

\bibitem[Wang et~al.(2021{\natexlab{b}})Wang, Chen, and Chen]{wang2021egat}
Ziming Wang, Jun Chen, and Haopeng Chen.
\newblock Egat: Edge-featured graph attention network.
\newblock In \emph{International Conference on Artificial Neural Networks},
  pages 253--264. Springer, 2021{\natexlab{b}}.

\bibitem[Li et~al.(2017)Li, Yu, Shahabi, and Liu]{li2017diffusion}
Yaguang Li, Rose Yu, Cyrus Shahabi, and Yan Liu.
\newblock Diffusion convolutional recurrent neural network: Data-driven traffic
  forecasting.
\newblock \emph{arXiv preprint arXiv:1707.01926}, 2017.

\bibitem[Steentoft et~al.(2024)Steentoft, Lee, and
  Schl{\"a}pfer]{steentoft2024quantifying}
Aike Steentoft, Bu-Sung Lee, and Markus Schl{\"a}pfer.
\newblock Quantifying the uncertainty of mobility flow predictions using
  gaussian processes.
\newblock \emph{Transportation}, 51\penalty0 (6):\penalty0 2301--2322, 2024.

\bibitem[Huang et~al.(2023)Huang, Jin, Candes, and
  Leskovec]{huang2023uncertainty}
Kexin Huang, Ying Jin, Emmanuel Candes, and Jure Leskovec.
\newblock Uncertainty quantification over graph with conformalized graph neural
  networks.
\newblock \emph{Advances in Neural Information Processing Systems},
  36:\penalty0 26699--26721, 2023.

\bibitem[Watson et~al.(2021)Watson, Lin, Klink, Pajarinen, and
  Peters]{watson2021latent}
Joe Watson, Jihao~Andreas Lin, Pascal Klink, Joni Pajarinen, and Jan Peters.
\newblock Latent derivative bayesian last layer networks.
\newblock In \emph{International Conference on Artificial Intelligence and
  Statistics}, pages 1198--1206. PMLR, 2021.

\bibitem[Rodrigues and Pereira(2020)]{rodrigues2020beyond}
Filipe Rodrigues and Francisco~C Pereira.
\newblock Beyond expectation: Deep joint mean and quantile regression for
  spatiotemporal problems.
\newblock \emph{IEEE transactions on neural networks and learning systems},
  31\penalty0 (12):\penalty0 5377--5389, 2020.

\bibitem[Yu and Moyeed(2001)]{yu2001bayesian}
Keming Yu and Rana~A Moyeed.
\newblock Bayesian quantile regression.
\newblock \emph{Statistics \& Probability Letters}, 54\penalty0 (4):\penalty0
  437--447, 2001.

\bibitem[Schulman et~al.(2017)Schulman, Wolski, Dhariwal, Radford, and
  Klimov]{schulman2017proximal}
John Schulman, Filip Wolski, Prafulla Dhariwal, Alec Radford, and Oleg Klimov.
\newblock Proximal policy optimization algorithms.
\newblock \emph{arXiv preprint arXiv:1707.06347}, 2017.

\end{thebibliography}

\appendix
\section*{Appendix}

\begin{figure}[H]
\centering
\includegraphics[width=\linewidth]{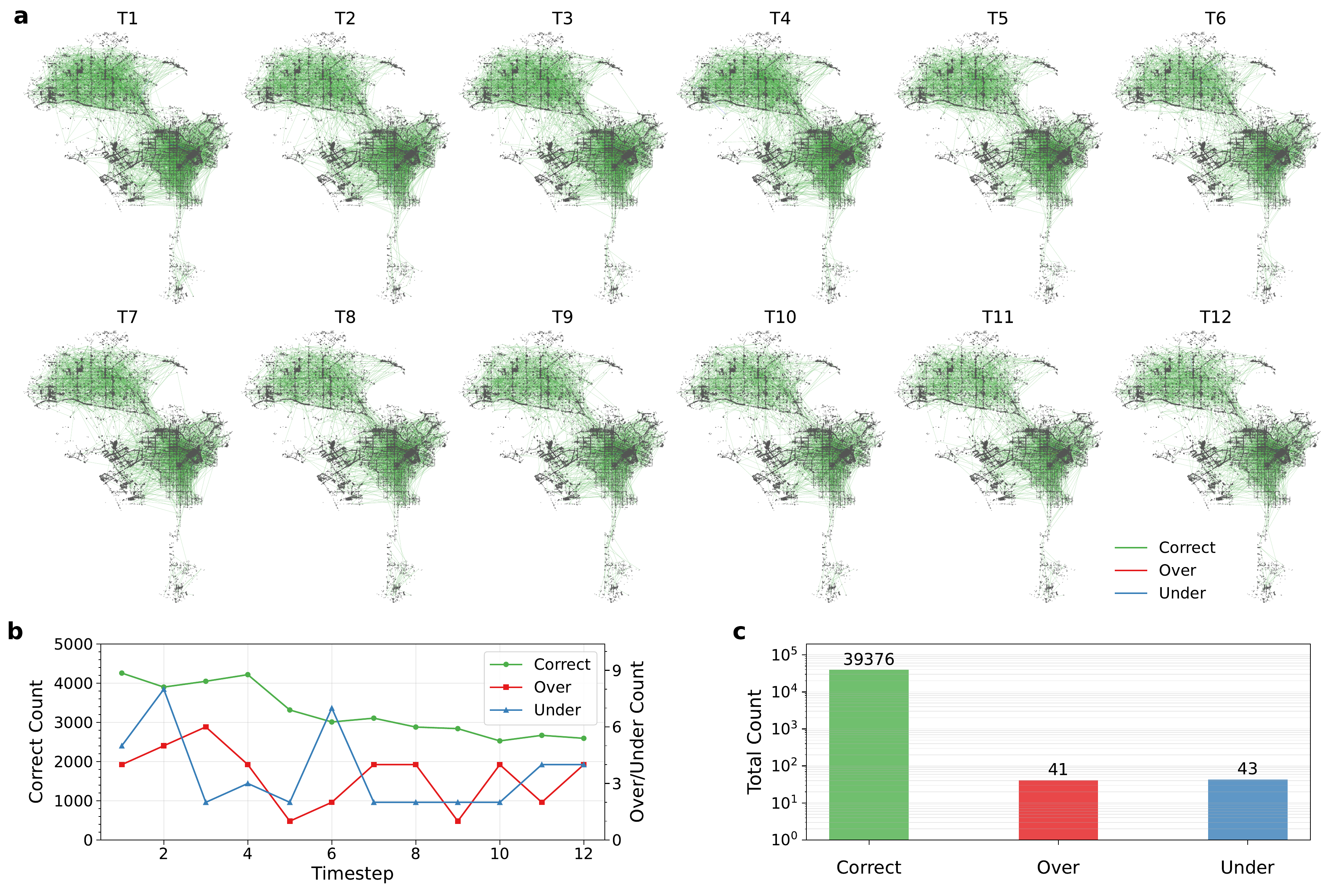}
\caption*{\textbf{Extended Data Fig. 1 \textbar{} Morning population flow predictions in the Los Angeles test dataset (07:00-10:00, March 29, 2023).} 
\textbf{a}, Spatial subgraphs showing edge-level prediction outcomes across consecutive 15-minute intervals during the morning period in Los Angeles.
Nodes represent points of interest with size of 38628. 
Edges are colored by prediction type: correct (green), overpredicted (red), and underpredicted (blue). 
\textbf{b}, Line plots summarizing the number of correct, overpredicted, and underpredicted edges at each 15-minute timestep. 
\textbf{c}, Total edge-level prediction counts over the 3-hour morning period, displayed on a logarithmic scale.}
\label{fig:prediction_LA_15m}
\end{figure}

\begin{figure}[H]
\centering
\includegraphics[width=\linewidth]{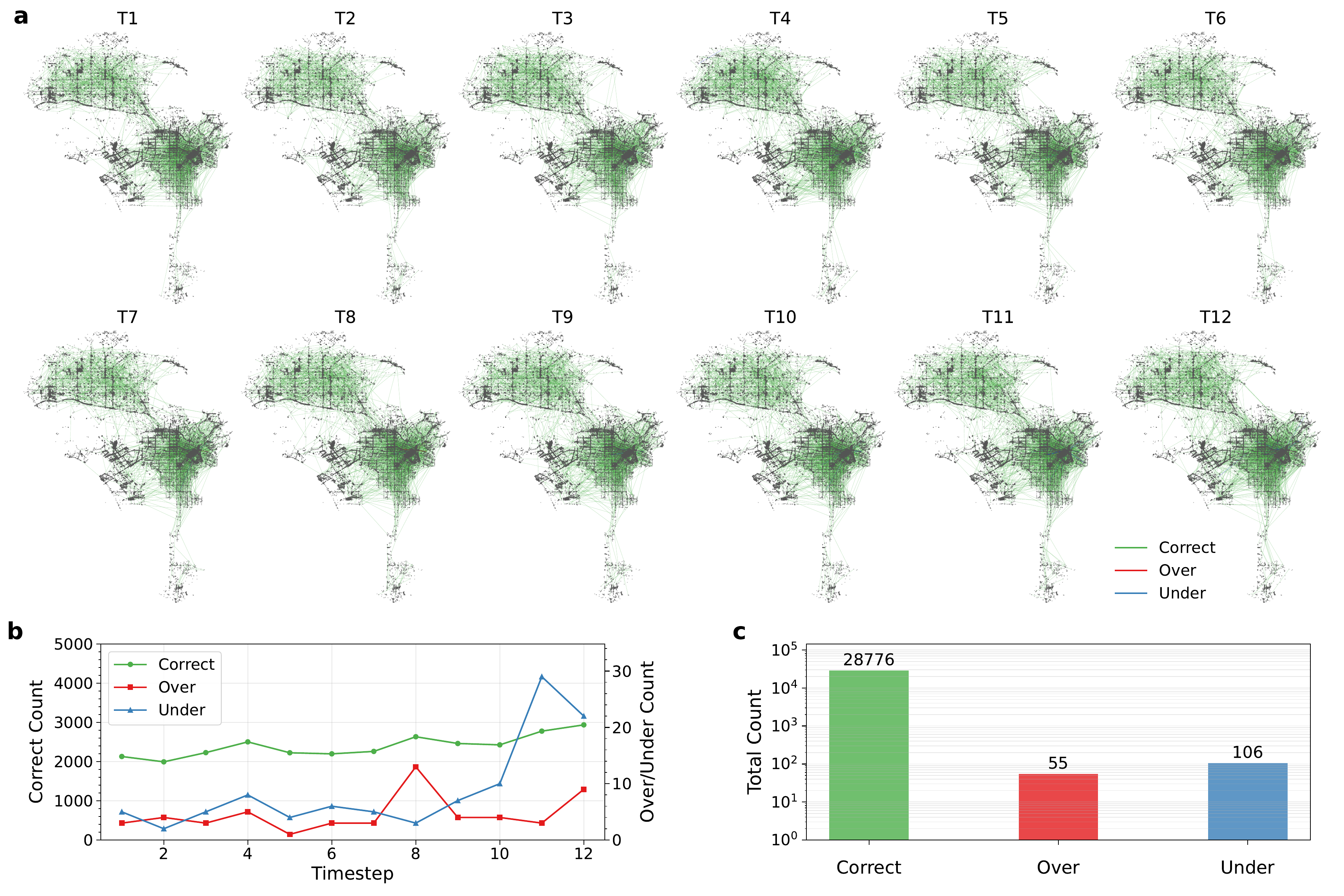}
\caption*{\textbf{Extended Data Fig. 2 \textbar{} Midday population flow predictions in the Los Angeles test dataset (12:00-15:00, March 29, 2023).} 
\textbf{a}, Spatial subgraphs showing edge-level prediction outcomes across consecutive 15-minute intervals during the midday period in Los Angeles.
Nodes represent points of interest with size of 38628. 
Edges are colored by prediction type: correct (green), overpredicted (red), and underpredicted (blue). 
\textbf{b}, Line plots summarizing the number of correct, overpredicted, and underpredicted edges at each 15-minute timestep. 
\textbf{c}, Total edge-level prediction counts over the 3-hour midday period, displayed on a logarithmic scale.}
\label{fig:prediction_LA_15m}
\end{figure}

\begin{figure}[H]
\centering
\includegraphics[width=\linewidth]{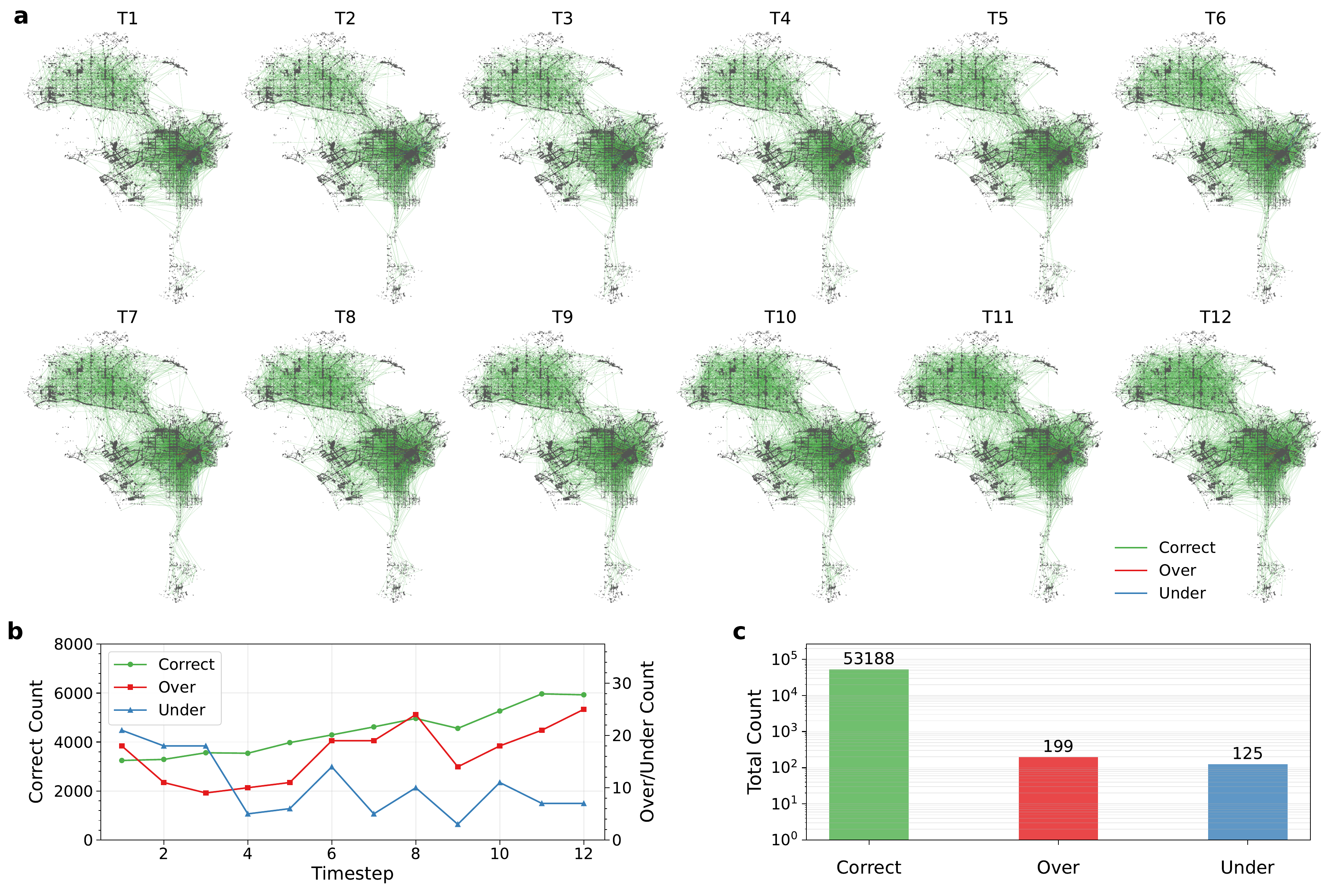}
\caption*{\textbf{Extended Data Fig. 3 \textbar{} Evening population flow predictions in the Los Angeles test dataset. (17:00-20:00, March 29, 2023).} 
\textbf{a}, Spatial subgraphs showing edge-level prediction outcomes across consecutive 15-minute intervals during the evening period in Los Angeles.
Nodes represent points of interest with size of 38628. 
Edges are colored by prediction type: correct (green), overpredicted (red), and underpredicted (blue). 
\textbf{b}, Line plots summarizing the number of correct, overpredicted, and underpredicted edges at each 15-minute timestep. 
\textbf{c},Total edge-level prediction counts over the 3-hour evening period, displayed on a logarithmic scale.}
\label{fig:prediction_LA_15m}
\end{figure}

\begin{figure}[H]
\centering
\includegraphics[width=\linewidth]{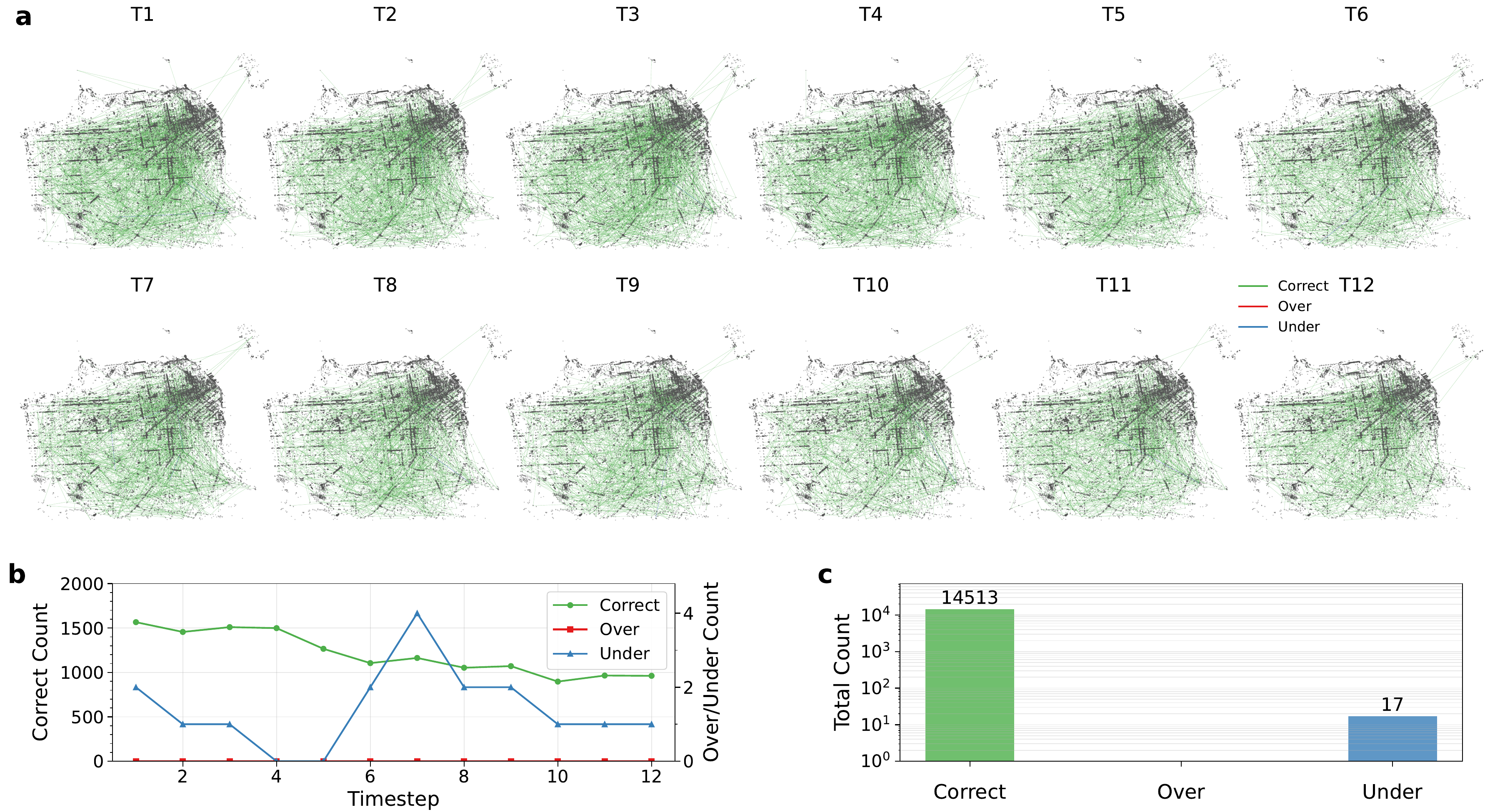}
\caption*{\textbf{Extended Data Fig. 4 \textbar{} Morning population flow predictions in the San Francisco test dataset (07:00-10:00, March 29, 2023).} 
\textbf{a}, Spatial subgraphs showing edge-level prediction outcomes across consecutive 15-minute intervals during the morning period in San Francisco.
Nodes represent points of interest with size of 28580. 
Edges are colored by prediction type: correct (green), overpredicted (red), and underpredicted (blue). 
\textbf{b}, Line plots summarizing the number of correct, overpredicted, and underpredicted edges at each 15-minute timestep. 
\textbf{c}, Total edge-level prediction counts over the 3-hour morning period, displayed on a logarithmic scale.}
\label{fig:prediction_SF_15m}
\end{figure}

\begin{figure}[H]
\centering
\includegraphics[width=\linewidth]{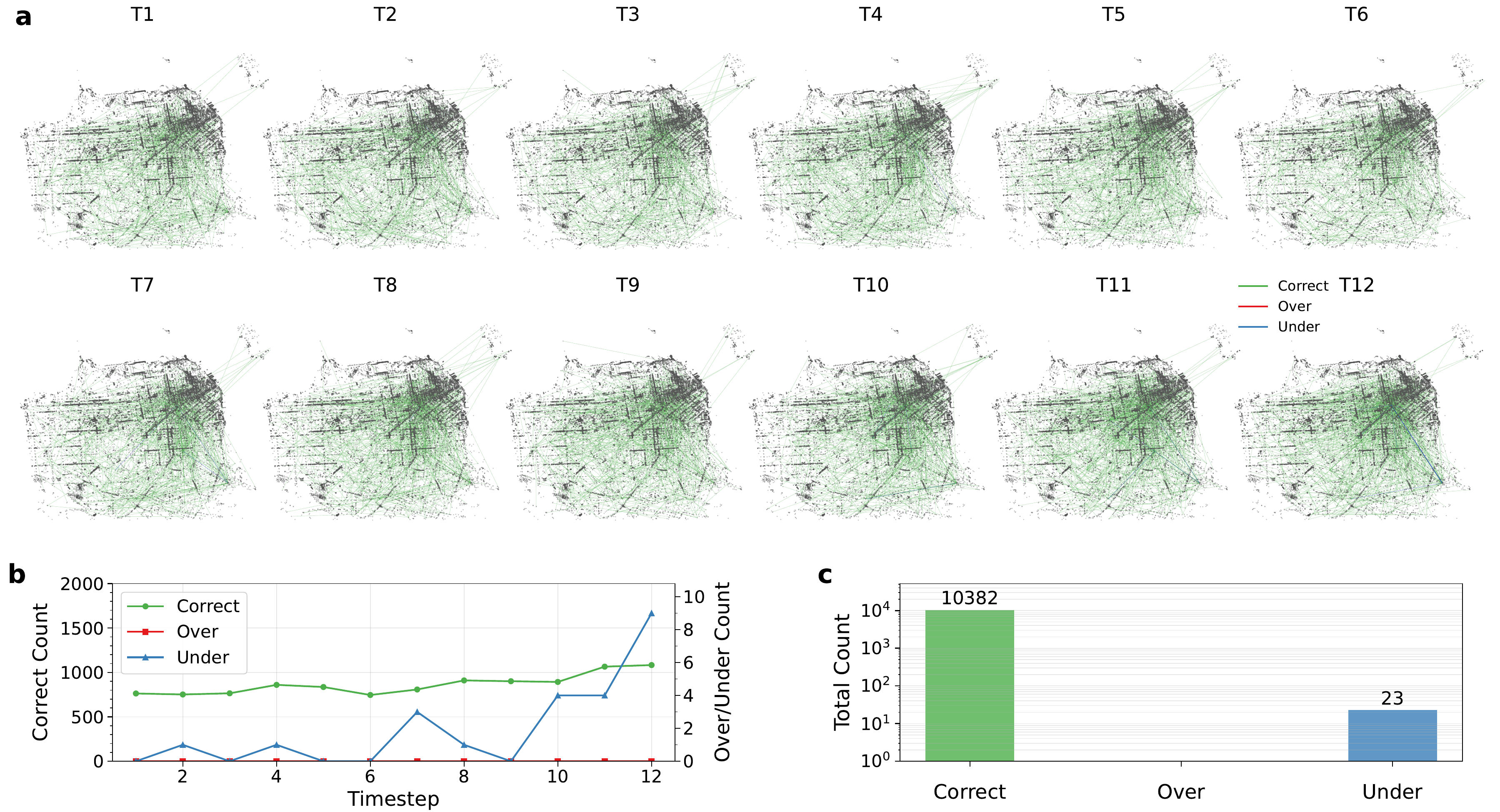}
\caption*{\textbf{Extended Data Fig. 5 \textbar{} Midday population flow predictions in the San Francisco test dataset (12:00-15:00, March 29, 2023).} 
\textbf{a}, Spatial subgraphs showing edge-level prediction outcomes across consecutive 15-minute intervals during the midday period in San Francisco.
Nodes represent points of interest with size of 28580. 
Edges are colored by prediction type: correct (green), overpredicted (red), and underpredicted (blue). 
\textbf{b}, Line plots summarizing the number of correct, overpredicted, and underpredicted edges at each 15-minute timestep. 
\textbf{c}, Total edge-level prediction counts over the 3-hour midday period, displayed on a logarithmic scale.}
\label{fig:prediction_SF_15m}
\end{figure}

\begin{figure}[H]
\centering
\includegraphics[width=\linewidth]{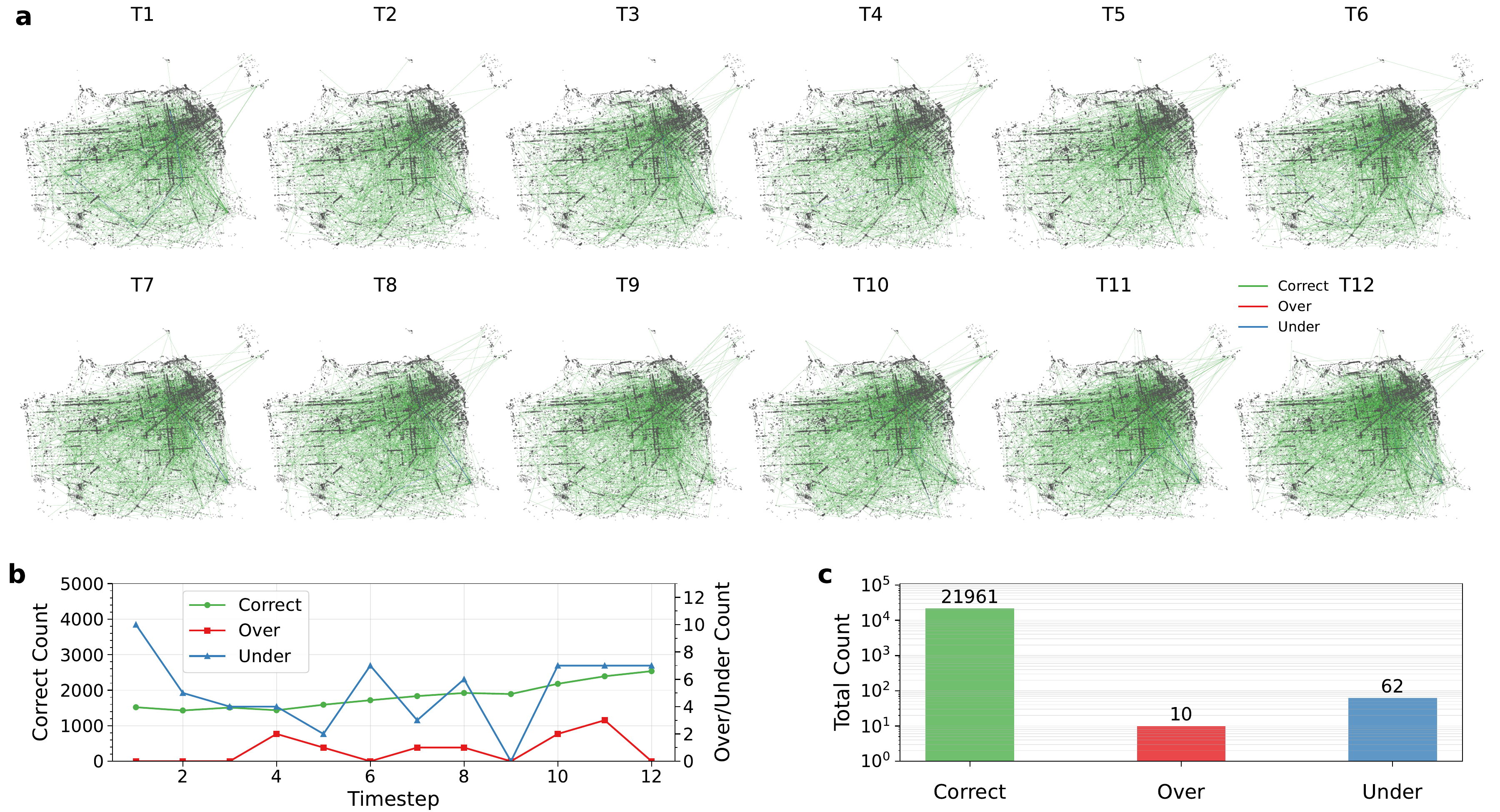}
\caption*{\textbf{Extended Data Fig. 6 \textbar{} Evening population flow predictions in the San Francisco test dataset. (17:00-20:00, March 29, 2023).} 
\textbf{a}, Spatial subgraphs showing edge-level prediction outcomes across consecutive 15-minute intervals during the evening period in San Francisco.
Nodes represent points of interest with size of 28580. 
Edges are colored by prediction type: correct (green), overpredicted (red), and underpredicted (blue). 
\textbf{b}, Line plots summarizing the number of correct, overpredicted, and underpredicted edges at each 15-minute timestep. 
\textbf{c},Total edge-level prediction counts over the 3-hour evening period, displayed on a logarithmic scale.}
\label{fig:prediction_SF_15m}
\end{figure}

\end{document}